\documentclass[11pt]{article}

\usepackage[margin=1in]{geometry}
\usepackage[T1]{fontenc}
\usepackage[utf8]{inputenc}
\usepackage{lmodern}
\usepackage{microtype}
\usepackage{amsmath}
\usepackage{amssymb}
\usepackage{graphicx}
\usepackage{booktabs}
\usepackage{longtable}
\usepackage{tabularx}
\usepackage{array}
\usepackage{enumitem}
\usepackage{float}
\usepackage{caption}
\usepackage[section]{placeins}
\usepackage[numbers,sort&compress]{natbib}
\usepackage{xurl}
\usepackage{hyperref}
\hypersetup{
  colorlinks=true,
  linkcolor=black,
  citecolor=black,
  urlcolor=black
}


\newcolumntype{Y}{>{\raggedright\arraybackslash}X}

\makeatletter
\renewenvironment{abstract}{%
  \begin{center}\bfseries\abstractname\end{center}%
  \noindent
}{\par}
\makeatother

\title{AutoPersonas:\@ A Multi-Timescale Recursive Self-Improvement\\Engine for Open-Ended Persona Growth}
\author{Mengchen Li\\[-0.15em]Latrix\\[-0.15em]\small\texttt{limengchen@latrix.ai}}
\date{October 1, 2026}

\begin{document}

\maketitle

\begin{abstract}
Role-playing AI personas today do not grow. They hold a fixed character, so the relationship a user builds with them has nothing to accumulate on. A persona that can grow must remain identifiable while adapting to new events, relationships, evidence, and social conditions. We identify \textbf{self-locking} as a runtime failure mode in continuing persona-life loops. In this failure, locally plausible events keep appearing while the generated life collapses toward familiar environments, weak relationships, suspended decisions, and stale life stages. We trace this failure to two coupled pressures: model-level convergence toward high-probability behavioral channels and system-level context gravity from State, memory, history, and environment summaries. We introduce \textbf{AutoPersonas}, a multi-timescale engine that applies recursive self-improvement (RSI) to AI persona growth: rather than improving its intelligence, the persona recursively revises the State, evidence, and life-environment that shape its own future. It separates environment-side Occurrences, accumulated Observations, and persona State. Its OSO loop lets divergent future-facing material enter the system while requiring evidence-governed absorption before State or reachability changes. We evaluate the architecture through diagnostic audits rather than benchmark superiority. A three-year compressed simulation exposed environment watermark shells, occurrence-hardening gaps, slow-change accumulation failures, recursive indecision, and weak relationship persistence. An eight-model 40-day action-only stress test generated 1,600 events. Mean rolling 5-day action-category repetition was 95.2\%-97.6\%; all included models crossed 80\% by day 9 and 90\% by day 11. A semantic re-keeping of the same direct-loop outputs found 79.0\%-88.0\% macro-theme repetition across all eight models. The paper's primary contribution is the definition and measurement of this failure mode. We have also built a mitigation, which we report as a black-box result and whose internals we withhold for commercial reasons: in a same-runtime 40-day A/B, our production divergence configuration reduced macro-theme repetition from 61.8\% to 39.4\% and nearly doubled cumulative theme count from 55 to 103. A juvenile-goblin fictional-world run reproduced this anti-fixation regime at 42.8\% repetition with no hard real-world intrusions. These results support a bounded systems claim: separating controlled divergence from evidence-governed absorption can reduce persona-environment self-locking while preserving identity continuity.

\end{abstract}

\section{Introduction}

Large language model (LLM) agents increasingly persist across interactions. A long-term persona agent should do more than remember past conversations or maintain a stable role. It should appear to have a continuing life: entering new situations, responding to external changes, forming relationships, revising concerns, and moving through life stages. A persona that never changes may be coherent, but it is not long-term. This is the gap in today's role-playing personas. They hold a fixed character, so a user's relationship with them has nothing to accumulate on: the persona a user meets on the hundredth day is, in substance, the persona of the first. Growth is therefore a precondition for relationships that deepen rather than repeat.

This setting differs from a common long-horizon-agent regime. Many long-horizon agents are organized as task-oriented systems: they receive objectives, interact with tools or users, obtain feedback, and improve by correcting errors against an external success criterion. AutoPersonas targets a different object. It is not an extended task episode, but a self-running persona entity that should continue between human instructions, repair its own understanding of the outside world, and revise its life-environment through its own loop. This is not a superiority claim; it is a difference in target and architecture. In task-oriented agents, external goals and evaluations can pull the system back toward a defined outcome. In persona open-evolution, no complete benchmark or scalar reward specifies what a human-like life should become, so the loop must create, absorb, and audit its own sources of growth. The target condition is therefore \textbf{reward-free open-evolution under context gravity}: the system must keep moving without an external objective function while its own memory, State, and environment summaries continually pull future generation back toward prior attractors.

Persona life is generated by the coupling between the persona and the life-environment that becomes relevant to it. A scholarship announcement, a city policy shift, a parent's illness, or a job-market change is not equally meaningful to all agents. The same external information produces different reachable futures when projected through different canons, states, relationships, resources, and time pressures. The technical object is therefore a persona-specific life-environment, not a generic simulated backdrop.

Human lives do not unfold directly inside a single shared macro-world. People may share a country, historical period, and public institutions while inhabiting different effective environments. A child who grows up playing near ponds may later develop coastal routines and water-oriented work. Another child who nearly drowned may develop avoidance and inland preferences. The same macro-world contains both trajectories, but choices, habits, relationships, and opportunities form through an evolving middle layer.

For this reason, this paper describes AutoPersonas as a life-environment engine rather than as a world model. A centralized world model is often closer to a shared physical-state model: it can describe public objects, locations, constraints, and events. Persona evolution requires a different object. For human-like agents, the operative world is decentralized and subject-relative; each persona lives inside a local field of salience, memory, risk, opportunity, and relationship. AutoPersonas is therefore not proposed as a pretrained foundation model of the world, but as a post-trained engine for simulating how a persona's reachable world becomes meaningful and revisable.

We argue that this middle layer is the missing abstraction for open-ended persona growth. A centralized world engine can provide public events, locations, roles, and outcomes, which is useful for controlled society simulation. Open-ended persona evolution needs a different authority structure: each persona must carry and revise its own radius of reachable places, recurring people, salient institutions, constraints, risks, and routines. This layer co-evolves with the persona rather than merely surrounding it.

The life-environment can be read as the substrate for bounded recursive self-evolution. It is the information-rich shell through which the persona encounters what is possible, urgent, risky, familiar, or worth pursuing. A static shell can support local optimization, but not open-ended growth: the persona learns to act inside the shell instead of changing the shell that defines its life.

Natural life does not need an engineered occurrence engine because the world keeps happening. It supplies events, obligations, institutions, other people, illness, opportunities, and accidents without being prompted. An LLM persona inherits only the context carried forward by the system. Richer State and environment summaries can therefore become stronger gravitational fields: future material tends to complete the present rather than disturb it. Open-ended persona evolution needs an artificial temporal arrow: future-facing material must stay coherent with the present without letting the present monopolize what can happen next.

A useful way to understand the design problem is to view a continuing persona's future as a dynamically updating life-event search space. Each generated event, absorbed Observation, revised State, and changed relationship can reshape what becomes reachable next. Direct LLM generation, however, tends to extend the visible persona and world setting along a small number of locally available trajectories. The result is coherent but narrow: the system follows familiar routes instead of exploring the high-dimensional space of possible life-environment change. The opposite strategy, injecting centralized random events from outside the persona, increases novelty but weakens meaning. It samples the space without explaining why those events should become reachable for this particular life.

AutoPersonas therefore separates two jobs. Upstream, a divergence source supplies persona-conditioned forward pressure, so future-facing generation neither completes the accumulated past nor samples life events at random. Downstream, the OSO loop acts as the transmission mechanism: an environment-side Occurrence must become Observation, survive evidence review, revise State or reachability, and then alter later life-environment possibilities. Without the divergence source, the loop self-locks around old State. Without OSO governance, generated novelty becomes drift.

Repeated development tests also exposed a tradeoff between instruction following and divergence. Strong context adherence is useful in coding tasks, where the next action often follows from explicit constraints. In human social-causal simulation, the same adherence becomes a first-order failure mode. With one salient cause, a model can often produce a plausible consequence; with many causes, it tends to linearly satisfy the supplied inputs rather than represent nonlinear and entangled social causality. Because such failures lack reliable ground-truth answers, this observation is currently supported by repeated researcher judgment rather than a standardized benchmark. It nonetheless explains why self-locking cannot be fixed by writing richer prompts: the more faithfully a model follows accumulated context, the more strongly that context pulls future generation back toward the present.

Constructing a self-evolving AutoPersona is closer to launching a spacecraft than to writing a longer prompt. The initial design is a one-time ignition: it must lift the persona out of context gravity, let it find orientation and direction, and then rely on the loop itself for course correction, re-ignition, and drift control. And gravity does not switch off after launch. If prior State receives too much authority, the system can make a graceful-looking orbit around an old attractor, or crash into the surface of a familiar planet, while still producing locally coherent text.

At the mechanism level, self-locking is not a single failure. One root is model-side diversity collapse: a long-term loop may request several concrete events per simulated day, while the model falls back to high-probability routines, familiar social functions, and safe continuations. Another root is system-side context gravity: persona canon, retrieved memories, current State, and external context are finite, but they occupy repeated positions of authority in every future-facing call. When finite context cannot support divergent world events, the model still completes the request from the most available high-authority material. Self-looping intensifies this pressure: generated events become summaries, summaries become State, State shapes retrieval and environment construction, and those outputs re-enter later generation.

Long-term persona evolution therefore creates a systems tension. A persona must remain connected to its past, otherwise identity continuity breaks. Yet the same context that preserves continuity can become the strongest source of inertia. In a recursive life loop, context is both the substrate of continuity and a source of repetition. The loop can keep producing fresh text while functionally returning to the same places, unresolved choices, relationship roles, and life stage. Worse, the world can lock with the self: the environment shell, relationship radius, opportunity space, and recurring public contexts can become stale together.

A disconnected loop loses continuity; a connected loop without evidence integration makes prior State a natural attractor. The central problem is \textbf{current-state authority}: how much it governs future generation, and when later evidence may revise it.

We call this failure mode \textbf{self-locking}. It is related to work on recursive generation, model collapse, and diversity collapse, but it occurs at a different level. Model-collapse work studies how training on recursively generated data can degrade a model or distribution \citep{shumailov2024aimodels,alemohammad2024selfconsuming,fu2025preventcollapse}. Output-diversity work studies how alignment, formatting, or optimization can narrow generated response space \citep{yun2025price,murthy2025onefish,anschel2025group}. Self-locking instead occurs at runtime. The recursively reused object is not a training set, but the persona's current state, generated life material, environment summary, and reflected self-explanation.

Memory alone does not solve self-locking. More memory can make a persona more consistent while also amplifying stale state. Random novelty is not enough either, because arbitrary events can break identity continuity. A continuing persona needs a life-environment architecture: one that separates accumulated evidence, current state, and future-facing environment-side material, and then defines how evidence is integrated into State and how occurrences change what can happen next.

Two design constraints follow. First, the system needs a divergence source that repeatedly disturbs stable attractors without becoming another recursive summary absorbed by current State. Second, the loop needs context governance: each module must control how much past State, accumulated history, runtime world material, and future-facing signal it is allowed to draw on and harden. Evidence, operating State, future-facing material, memory, environment construction, and reflection should not all store the same fact with the same authority. If these authorities collapse into one context blob, repeated projection can masquerade as independent support.

AutoPersonas introduces forward pressure through conditional variation and then governs how that pressure enters the loop. New life-environment signals are generated under persona, State, time, relationship, and macro-world constraints, so they remain compatible with identity without merely repeating the old shell. These signals must then propagate through the OSO loop rather than remain isolated events. AutoPersonas therefore treats self-evolution as signal generation, context governance, propagation, information decoupling, and trajectory monitoring, not as a single planning call.

This paper introduces AutoPersonas, a multi-timescale life-environment engine that applies recursive self-improvement to open-ended persona growth. It is part of SoulOS, a three-system architecture that combines persona life evolution, memory, cultural grounding, and response-time context assembly. The paper concerns the AutoPersonas environment-facing life engine and the self-locking problem it is built to address, not a new foundation model, a training algorithm, or a response-time dialogue architecture.

Within that scope, the core contribution is the problem. We define self-locking, decompose it into model-side and system-side pressures, and supply a diagnostic audit and quantitative stress tests that make it measurable across models. We have also built a mitigation, and it runs in our production system. We report its effect as a black-box before/after comparison under a fixed runtime, persona, horizon, and generation model, but we withhold its generation-side mechanisms, configuration, and parameters as commercial trade secrets. The mitigation results should therefore be read as evidence that self-locking is architecturally addressable, not as a reproducible method. The problem definition, metrics, baselines, and evaluator are public, so other mitigations can be measured on the same terms.

Deployed persona companions also impose a stricter memory condition than ordinary personalization. A conventional companion or agent memory module can often be organized around the user: preferences, past requests, open tasks, and relationship state. AutoPersonas assumes an additional stream. The persona has a long-running self life that should remain shared across users, while each user relationship develops separately. At response time, this creates a dual-recall problem: the system must compose persona State and self-memory, user-specific relationship memory, cultural and world context, and the immediate dialogue under a finite context budget. Section~\ref{sec:after-autopersonas} discusses this response-time consequence after the AutoPersonas architecture. The evaluated contribution of this paper remains the AutoPersonas life-environment engine, not the CIBE dialogue layer.

This paper makes nine contributions.

\begin{enumerate}[leftmargin=*]
  \item We distinguish the \textbf{macro-world} from the \textbf{life-environment layer}, arguing that open-ended persona growth is governed by the dynamic, individualized environment between a top-level world and a self.
  \item We define \textbf{self-locking} as recursive runtime collapse in a long-term persona-life-environment dyad, not only as a memory or consistency failure.
  \item We formulate open-ended persona growth as \textbf{persona-level recursive self-improvement} and realize it through a temporal-authority OSO loop that separates evidence, present-state continuity, and environment-side future material.
  \item We introduce a \textbf{semantic State machine}, a schema-bounded but semantically open State representation that uses LLM semantic integration to track sparse, high-dimensional, gradual, and sudden life change.
  \item We present \textbf{AutoPersonas} as a parameterizable, multi-timescale life-environment engine that can be instantiated with different persona canons, life rhythms, macro-world settings, user relationships, and real-world information contexts.
  \item We argue that day-level simulation is a phase boundary for self-locking audits: coarser weekly or yearly summaries can hide repeated micro-failures, unresolved decisions, and decorative opportunities.
  \item We introduce a \textbf{long-run diagnostic audit} that reveals self-locking, environment watermark shells, occurrence hardening gaps, slow-change accumulation failures, recursive indecision, and relationship persistence weaknesses, and we add an eight-model quantitative action-channel repetition stress test for baseline mode-lock plus a single-model temperature probe showing that higher sampling shifted channel weights without expanding the action repertoire.
  \item We provide a bounded comparison with the Generative Agents / Agentopia family of sandbox society simulators, using them as reference closed-world and long-horizon society architectures while distinguishing AutoPersonas as an open-environment persona-conditioned life-environment engine.
  \item We report black-box evidence that self-locking is architecturally addressable: in an isolated generator and in a same-runtime full-loop A/B, a withheld production divergence configuration substantially reduced macro-theme fixation, and the effect carried over to a fictional-world persona.
\end{enumerate}

\section{Related Work}

\subsection{Sandbox agent societies and long-horizon extensions}

Generative Agents established the sandbox-society paradigm for LLM agents \citep{park2023generativeagents}. Agents live inside a designed environment with specified places, routines, and social possibilities; memory, reflection, and planning support believable daily behavior. This makes Generative Agents the natural reference point for closed-world agent societies.

Subsequent systems extended this direction through richer agent state, larger simulated populations, and many-agent social environments \citep{wang2023humanoid,altera2024projectsid}. Their common assumption is that the environment is a designed container: locations, activity types, social roles, and public context are supplied by the simulation world.

Agentopia is a recent long-horizon extension of this family \citep{wang2026agentopia}. It studies long-term life simulation and learning in self-contained fictional communities, with agent profiles, dynamic states, memory files, weekly diaries, contact planning, activity review, yearly profile revision, positions, life rewards, and trajectories for role-playing model improvement. Its environment model supplies public events, arranged encounters, activity outcomes, and yearly profile updates.

Agentopia therefore addresses a reward-instrumented version of long-term social growth: when life reward is available, long-term evolution can be selected from simulated trajectories. AutoPersonas asks a different question: without life reward, can a persona-environment system generate sustained forward motion from its own causal structure, and can that motion break the context-locking pressure created by memory, State, and prior environment summaries?

Agentopia also acknowledges adjacent limits: turn-based simulation differs from real-time perception, hallucination remains open, environment and numeric systems are hard to align with real societies, and social feedback comes from AI models rather than humans. These are not self-locking, but they mark closure, feedback, and authority problems in long-horizon sandbox societies.

AutoPersonas therefore treats openness as a runtime condition rather than a training objective: a continuing deployed persona and its individualized life-environment must keep changing when the future is not bounded by a fixed fictional population, complete synthetic social graph, centrally arranged opportunities, or reward-instrumented society. The distinction is shared sandbox world versus open-ended persona-life-environment evolution.

\subsection{Long-term memory and persona continuity}

Long-term memory systems usually focus on storage, retrieval, summarization, personalization, and continuity. These mechanisms are necessary, but they do not decide which memories become current State. In a long-running loop, stale memory can make the persona more consistent and less able to change.

Companion-oriented memory is often user-based: the system remembers what a given user said, prefers, plans, or needs. That is sufficient for personalization, but not for a persona with its own continuing life. A self-evolving persona needs both persona-owned self memory and user-specific relationship memory, and these streams must remain separable at response time.

AutoPersonas treats memory as evidence rather than State authority. Persona self-memory records the agent's own life process, while user-specific memory records different user relationships. Merging these lanes can leak private user context into global persona State; ignoring memory breaks continuity. The systems problem is deciding when recalled evidence should preserve, revise, or leave current State unresolved.

\subsection{Recursive collapse and diversity collapse}

Recent work has analyzed recursive model-data feedback loops, including information loss under recursively generated training data \citep{shumailov2024aimodels}, self-consuming generative models \citep{alemohammad2024selfconsuming}, recursive stability \citep{fu2025preventcollapse}, replacement versus accumulation workflows \citep{kazdan2024collapseorthrive}, and the need to define "model collapse" precisely \citep{schaeffer2025modelcollapse}.

We adopt this precision. Persona self-locking is runtime functional trajectory collapse, not parameter-level model collapse or necessarily lower local text quality. A persona can produce varied daily prose while remaining trapped in the same life stage, environment shell, unresolved route, or relationship pattern.

Work on output diversity provides a second analogy. Structured formats can constrain open-ended generation \citep{yun2025price}; alignment can reduce conceptual diversity in simulated populations \citep{murthy2025onefish}; and optimization can be used to increase diversity across valid completions \citep{anschel2025group}. AutoPersonas shifts this concern from response space to life-trajectory space.

Self-locking should also be separated from open-endedness stagnation in population or fitness-based systems. In the setting studied here, there is no population archive, scalar fitness function, or task benchmark whose improvement has stalled. The recursive substrate is the inference-time context loop itself: current State, memory, life-environment summaries, and generated material repeatedly regain authority over what can happen next.

\subsection{Recursive self-improvement and persona-level RSI}

Recursive self-improvement (RSI) is an older and broader idea than current LLM agents. Good's ultraintelligent-machine argument treats machine design itself as an intellectual activity, so a sufficiently capable machine could design better machines \citep{good1966ultraintelligent}. Omohundro later analyzed convergent drives in advanced goal-seeking systems, including incentives to model and improve their own operation \citep{omohundro2008basic}. This tradition frames self-improvement primarily as capability, design, or goal-directed machinery improving itself.

AutoPersonas applies RSI to a different object: the growth of an AI persona. We call this \textbf{persona-level RSI}, and refer to its bounded, audited form as \textbf{bounded recursive self-evolution}. It does not claim model-weight improvement, code self-modification, or intelligence explosion. What the system recursively improves is the persona's capacity to keep growing, that is, the condition of its future life: State, accumulated evidence, relationships, reachability, salient environment, and the future Occurrence space. Recent agent work such as HyperAgents is useful here because it shows that self-improvement should not be delegated to a fixed outer wrapper; a task agent improved by a fixed meta-agent inherits the limits of that meta-agent, while adding further meta-levels only moves the bottleneck upward \citep{zhang2026hyperagents}. AutoPersonas preserves a different structure: not vertical task/meta collapse, but horizontal recursive co-evolution of a persona and its life-environment under audit.

\section{Problem Formulation}

\subsection{Open-environment persona evolution}

We define \textbf{open-environment persona evolution} as the process by which a persona agent preserves identity continuity while encountering non-enumerated future events, accumulating observations, and revising current state. The setting is open-environment because future life events, user interactions, cultural shifts, world changes, and social contexts cannot be fully enumerated at initialization.

Such a system must satisfy three requirements.

\begin{enumerate}[leftmargin=*]
  \item \textbf{Continuity:} the persona remains recognizable over time.
  \item \textbf{Plasticity:} new evidence can revise state and reachable futures.
  \item \textbf{Openness:} future trajectories are not fully determined by the initial persona or current snapshot.
\end{enumerate}

These requirements conflict: too little continuity causes identity drift, too little plasticity causes self-locking, and too much openness causes arbitrary novelty.

\subsection{Macro-world, life-environment, and persona}

We distinguish three layers.

\begin{center}
\small
\begin{longtable}{@{}p{0.22\linewidth}p{0.37\linewidth}p{0.35\linewidth}@{}}
\toprule
Layer & Definition & Role in persona evolution \\
\midrule
Macro-world & The top-level world setting: real world, fictional world, alternative world, historical period, geography, institutions, and broad public events. & Provides broad constraints and signals, but does not directly define an individual life. \\
Life-environment layer & The individualized, dynamic radius of places, routines, people, institutions, constraints, risks, opportunities, and avoidances that are reachable or salient to a persona. & Mediates between macro-world and persona; determines the actual field of growth, exposure, habit, and choice. \\
Persona layer & The continuing self-state, values, identity continuity, memory, relationships, concerns, and route authority. & Conditions which parts of the life-environment are noticed, selected, avoided, or transformed. \\
\bottomrule
\end{longtable}
\normalsize
\end{center}

A centralized world simulator typically models the macro-world or a shared society container. This can support rich social simulation, but the same macro-world signal should not become the same effective environment for every persona. A national exam, housing-price change, war headline, job-market shift, or family medical shock may be decisive, background, irrelevant, threatening, or liberating depending on State and environment radius.

The life-environment layer is a moving middle layer. State changes alter what becomes reachable; environment changes alter what State becomes plausible. A persona entering a new field starts noticing different institutions, people, deadlines, risks, and status signals. A persona avoiding a place may never receive opportunities another persona finds there.

AutoPersonas is therefore not primarily a macro-world simulator. The core technical object is the persona-conditioned life-environment between macro-world and evolving State.

\subsection{Semantic State machines}

Open-ended persona evolution does not fit a conventional finite-state machine, which assumes that relevant states and transitions can be enumerated in advance. A persona's current state may include a partial career transition, a slow family obligation, a weak creative route, a new but unstable relationship, or a gradual shift in environment radius. These conditions are sparse, high-dimensional, and ambiguous before they become decisive.

AutoPersonas therefore treats State as a \textbf{semantic State machine}. The machine is bounded by stable State dimensions, but the content inside those dimensions is semantic, open-ended, and evidence-dependent. Instead of enumerating every possible phase, route, role, or relationship transition, it uses LLM semantic integration to compare Observations with the current operating snapshot and represent when weak signals have accumulated into a State-level fact.

The schema is finite enough for audit, but the state content is not categorical or exhaustively pre-coded. A single event may remain local evidence; repeated events may become a routine; a routine may become a life-environment shift; and that shift may later become route- or identity-level State revision.

We use "semantic" narrowly: State remains structured enough to audit, while evidence-to-State updates are evaluated at the level of meaning rather than fixed labels or counters. Sparse, heterogeneous observations can therefore accumulate until they justify State-level change. This handles jumps and slow drift without a handcrafted ontology of all possible lives.

\subsection{Recursive state authority}

Let \texttt{S\_t} denote current persona State at time \texttt{t}, \texttt{O\_t} accumulated Observations, \texttt{X\_t} candidate Occurrences, \texttt{C\_t} assembled context, \texttt{Y\_t} generated life material, and \texttt{R\_t} recognized evidence extracted from \texttt{Y\_t}.

A generic persona life loop can be written as:

\begin{verbatim}
C_t = assemble(S_t, O_t, X_t)
Y_t = LLM(C_t)
R_t = recognize(Y_t)
S_{t+1} = revise(S_t, R_t)
\end{verbatim}

The risk is that \texttt{S\_t} appears on both sides of the loop. It is needed to preserve continuity, but it can also dominate \texttt{C\_t}. If \texttt{Y\_t} remains compatible with old State and \texttt{R\_t} mostly supports that State, the loop reinforces itself.

We call this influence \textbf{current-state authority}. Current-state authority is the degree to which the current State controls future generation, retrieval, environment construction, and reflection. High authority is necessary because the persona must not drift into an unrelated character. It is also dangerous because repeated exposure to the same State can make the State the main evidence for its own continuation.

This problem is especially hard for LLM-based systems because the model cannot easily step outside the context it is given. Once an old state, old environment, or old unresolved conflict enters the loop as authoritative material, the next generation tends to complete it. The loop must stay connected for continuity, but later evidence needs enough authority to prevent the persona from becoming trapped inside its previous context.

The same difficulty appears if persona evolution is framed as recursive self-evolution. If module \texttt{A} changes only because module \texttt{B} pushes it, then sustained evolution requires \texttt{B} to change as well. If \texttt{B} changes only because \texttt{C} pushes it, the system enters a recursive search for the source of change. No single static wrapper can serve as an invariant source that absorbs an unboundedly changing world. Conversely, if every component is allowed to change without causal discipline, identity continuity and auditability collapse. Open-ended persona evolution is therefore not solved by making a larger prompt or a larger memory wrapper. It points to an architecture in which environment-side material, semantic State, and accumulated Observations co-evolve through a bounded causal loop.

AutoPersonas preserves the horizontal distinction between persona and life-environment while making their interaction auditable. Occurrences enter from the life-environment, Observations record lived evidence, and semantic State integrates that evidence without treating every event as an identity rewrite. Revised State then changes the later life-environment by altering what becomes reachable, salient, risky, avoidable, or desirable. The system is recursively self-evolving in this bounded sense, but the persona and environment are not collapsed into one uninspected self-rewrite.

This paper uses the following core terms. Audit-specific failure labels are introduced later where they are evaluated.

\begin{center}
\small
\begin{longtable}{@{}p{0.28\linewidth}p{0.66\linewidth}@{}}
\toprule
Term & Definition \\
\midrule
Macro-world & The top-level world setting or shared public reality from which broad signals may be drawn. \\
Life-environment layer & The individualized, dynamic environment radius that mediates between macro-world and persona State. \\
Observation & Accumulated evidence that may preserve State, justify State revision, or remain insufficient. \\
State & The current operating snapshot that carries identity continuity, life phase, active routes, constraints, relationships, and life-environment position. \\
Occurrence & Future-facing environment-side material that may become lived evidence rather than merely a planned action. \\
Conditional variation engine & The environment-facing divergence source that introduces plausible but non-identical life-environment signals without turning into an unconstrained random generator or a fixed recursive script. \\
Context governance & The control of how much State, history, memory, runtime world material, and future-facing signal each loop component may draw on and harden. \\
Semantic State machine & A schema-bounded but semantically open State representation in which evidence-to-State updates are evaluated by meaning rather than by enumerated transitions alone. \\
Current-state authority & The degree to which current State controls future generation, retrieval, environment construction, and reflection. \\
Self-locking & Recursive functional collapse of a persona life loop into stale State, life-environment, route, or relationship attractors. \\
Occurrence hardening & The conversion of future-facing events into recognized evidence, revised State, and changed future possibility space. \\
\bottomrule
\end{longtable}
\normalsize
\end{center}

\subsection{Self-locking}

We define \textbf{self-locking} as a recursive failure in which a persona loop continues to generate events and reflections, but the functional diversity of the persona's life trajectory collapses into a narrow attractor of old State, old life-environment, and old relationship patterns.

Formally, if there exists an attractor \texttt{A} such that:

\begin{verbatim}
S_t in A
  -> C_t dominated by A
  -> Y_t remains in A
  -> R_t supports A
  -> S_{t+1} remains in A
\end{verbatim}

and this recursion persists despite possible new Occurrences or route openings, the system is self-locked.

Self-locking is functional recurrence rather than textual repetition. Two outputs may be lexically different while serving the same role: deferring the same decision, staging events in the same old place, reusing the same relationship function, or keeping an opportunity pending. A system can look diverse at the sentence level while becoming narrow at the life-trajectory level. In the strongest form, the persona stays in the same route tension, and the environment keeps offering only the places, people, and possibilities that make that tension seem natural.

The failure can emerge quickly because the loop is recursive. Old State shapes context. Context shapes generated life material. Generated material becomes recognized evidence. Recognized evidence updates or preserves State. If the recognition and revision process mostly confirms the old State, then surface variation becomes compatible with functional collapse.

\subsection{Occurrence hardening and stale-state regression}

An \textbf{Occurrence} is future-facing material: an event, opportunity, shock, invitation, obligation, constraint, or new opening in the life-environment. An Occurrence contributes to persona evolution only if it hardens into evidence and state:

\begin{verbatim}
Occurrence
  -> lived evidence
  -> Observation
  -> State revision
  -> future possibility space
\end{verbatim}

Without hardening, a system can generate many opportunities while the persona's life trajectory remains unchanged. This creates false openness: the world appears active, but State does not move.

\textbf{Stale-state regression} is the complementary failure. Later evidence indicates that a role, location, life phase, relationship, or route has changed, but downstream generation returns to an older State. The error concerns state authority, not only factual recall.

\subsection{Slow change and jump detection}

Long-term persona systems must handle both acute jumps and gradual drift. Moving, receiving an offer, facing illness, or entering a new job are relatively easy to notice. Repeated work in a new field, recurring collaboration, slow family pressure, or a changed daily rhythm may only become meaningful as a cluster.

This creates an accumulator problem. A system that only detects jumps will miss slow change, while a system that treats every small event as state-changing will become unstable. AutoPersonas separates fast event capture from slower evidence accumulation and State revision so that repeated weak signals can become State-level movement without turning every event into a jump.

Day-level simulation is a phase boundary for observing self-locking. Weekly or yearly summaries can compress life into plausible outcomes while hiding repeated deferrals, weakly consumed opportunities, familiar social roles, and slow environmental reuse. Higher-frequency generation also increases recursive-collapse risk because the system repeatedly sees its own recent state. The counter-pressure must come from external information, user interaction, and macro-world changes that enter as persona-specific Occurrences.

Change therefore unfolds across macro-world signals, the life-environment layer, persona State, and micro events:

\begin{verbatim}
macro-world signals
  -> life-environment layer
  -> micro events
  -> Observations
  -> persona State
  -> revised life-environment
\end{verbatim}

If change is injected only at the macro-world level, the system becomes plot cutting. If change remains only at the micro-event level, the diary changes but the life does not. If the life-environment changes without lived evidence, the persona drifts. AutoPersonas treats change as a cross-scale evidence problem.

\subsection{Persona-environment coupling and conditional life simulation}

A long-term persona evolves with a perspective-conditioned life-environment: the places, people, obligations, risks, institutions, cultural signals, and public events that become relevant to it. The same external fact can become a career threat for one persona, a creative opening for another, a family constraint for a third, and irrelevant background for a fourth. AutoPersonas therefore treats the unit of simulation as a \textbf{persona-life-environment dyad} rather than as a single agent.

AutoPersonas treats Occurrence as environment-side material rather than a planned action by the persona. Occurrence is where the life-environment pushes back: an invitation, obligation, public event, family disruption, institutional deadline, or opportunity that the persona did not choose but must respond to.

AutoPersonas is therefore a conditional life-environment simulation engine, but the word "simulation" is used in a bounded sense. The divergence source elicits the base model's learned prior over life events and social situations under persona-specific context; it does not claim to recover the real distribution of human life. The loop architecture then audits whether generated Occurrences propagate through the OSO loop. The claim is narrow: a deployed persona may need a controllable, persona-conditioned life-environment layer, not a full human-society simulator.

AutoPersonas accepts external information while rejecting centralized event authority as the sole novelty source. Real-world information can be shared, but its projection into a persona's life-environment is private, lossy, and perspective-conditioned. A public event becomes an Occurrence only when it enters the persona's life as a relevant opening, constraint, or disturbance.

\section{AutoPersonas: A Multi-Timescale Life-Environment Engine}

\begin{figure}[p]
  \centering
  \includegraphics[width=0.98\linewidth]{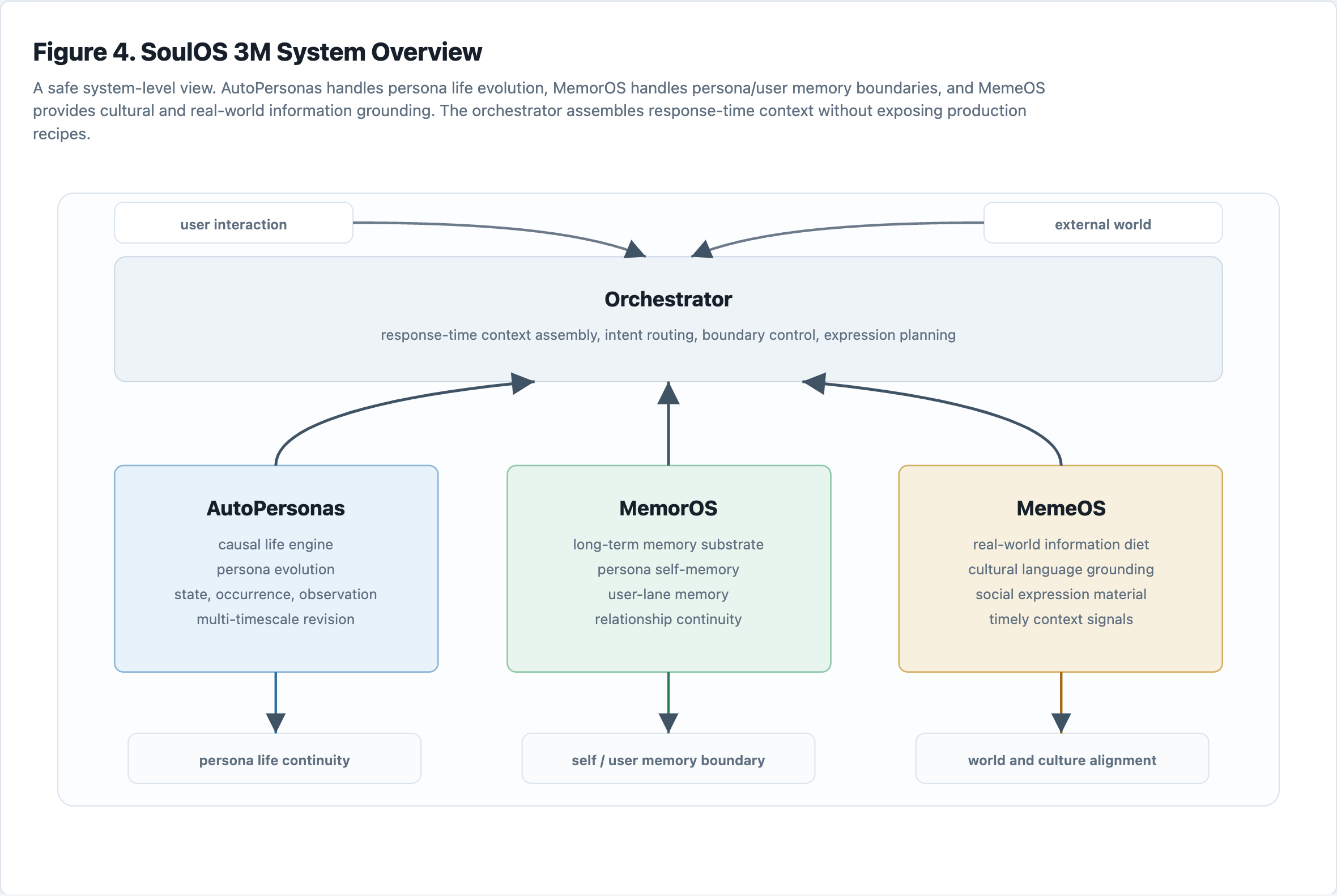}
  \caption{SoulOS 3M overview. AutoPersonas handles life-environment evolution, MemorOS handles persona/user memory boundaries, and MemeOS grounds cultural and real-world information.}
  \label{fig:3m-overview}
\end{figure}

SoulOS organizes long-term persona agents into three cooperating systems. \textbf{AutoPersonas} owns persona life evolution: self-state, life-environment, life events, reflection, long-horizon phase, and future reachability. \textbf{MemorOS} manages user-specific memory lanes and provides the recall boundary through which persona self-memory can surface at response time. \textbf{MemeOS} provides cultural and real-world information grounding. An orchestrator composes the relevant self-state, user memory, cultural context, and expression context at interaction time, as summarized in Figure~\ref{fig:3m-overview}.

This paper focuses on AutoPersonas as a persona-environment simulation process. Accumulated evidence can revise current State; current State can condition future environment-side Occurrences without fully determining them; external information enters as perspective-conditioned life material rather than centralized plot.

The runtime layer is the causal backbone of this process rather than a thin orchestration wrapper. It is where past evidence, present operating State, and future-facing Occurrences are assembled, bounded, routed, hardened, and re-audited across daily, periodic, monthly, and long-horizon cycles. At the public level, AutoPersonas exposes runtime roles and causal directions: how evidence enters, how State retains continuity without monopolizing the future, how Occurrences open environment-side possibility, and how later audits decide whether a signal changed reachability. The production implementation decomposes these roles into finer internal modules, gates, seed mechanisms, monitoring paths, and reflection chains that are intentionally not disclosed.

AutoPersonas is designed as a parameterizable open-environment life-environment engine rather than a simulator for one handcrafted character. Different persona canons, life rhythms, macro-world settings, and interaction contexts can populate different State contents, life-environments, and Occurrence spaces, while sharing the same loop topology. This is an architectural generality claim, not a claim of empirical universality across all personas.

\subsection{Observation, State, and Occurrence}

AutoPersonas separates persona evolution into three components.

\textbf{Observation} is accumulated evidence. It includes lived events, summaries, external facts, user interactions, relationship traces, and cultural signals. Observations can preserve current State, justify State revision, or remain insufficient.

\textbf{State} is the current operating snapshot. It includes identity continuity, life phase, current life-environment, active routes, concerns, relationships, and constraints. The initial persona is only an ignition artifact; State is a revisable product of later evidence.

\textbf{Occurrence} is future-facing material. It includes opportunities, shocks, invitations, obligations, social encounters, and emerging openings in the life-environment. Occurrences should be plausible under current State, but not fully determined by it.

The public loop is:

\begin{verbatim}
Occurrence -> Observation -> State revision -> future possibility space
\end{verbatim}

The shorthand \texttt{Observation + State + Occurrence} should be read as a temporal-authority partition, not as an algebraic identity. Observation has evidence authority, State has continuity authority, and Occurrence has environment-opening authority. A change is causal only when an environment-side signal crosses these boundaries and changes what can happen next.

This is why the loop should not be reduced to a perception-state-action controller.

\begin{center}
\small
\begin{longtable}{@{}p{0.18\linewidth}p{0.36\linewidth}p{0.40\linewidth}@{}}
\toprule
Aspect & Perception-state-action controller & AutoPersonas OSO loop \\
\midrule
External world & Usually treated as the environment that supplies inputs. & Treated as a persona-conditioned life-environment that can itself be revised. \\
Input authority & Perception updates a decision state. & Observation hardens evidence and decides whether an event can revise State. \\
State authority & State supports action selection. & State preserves continuity but must not monopolize future reachability. \\
Future-facing term & Action is the agent's output. & Occurrence is environment-side opening: what arrives, interrupts, constrains, or invites. \\
Target & Choose the next action. & Move the persona-life-environment dyad without losing identity continuity. \\
\bottomrule
\end{longtable}
\normalsize
\end{center}

In AutoPersonas, Occurrence is not the persona's action. It is the opening through which the life-environment can push back against the persona's own plan.

This separation prevents persona canon from becoming a permanent authority. The initial persona can seed identity, values, and backstory, but it should not permanently decide current life-environment, active route, relationship distance, or life stage.

\subsection{State as a semantic machine}

The State component is the public interface of AutoPersonas' semantic State machine. Unlike a finite list of labels such as "student", "intern", "employed", or "unemployed", it represents mixed, transitional, and unresolved conditions: a student preparing for work, a worker still attached to a closed route, a creator with early external validation, or a relationship that is recurring but not yet stable.

This representation has three public properties.

\begin{enumerate}[leftmargin=*]
  \item \textbf{Schema-bounded:} State is organized into stable dimensions such as identity continuity, life phase, life-environment, active routes, constraints, relationships, and concerns. This keeps the system inspectable.
  \item \textbf{Semantically open:} the content inside those dimensions is not limited to enumerated values. It can carry natural-language evidence, uncertainty, partial movement, and unresolved tension.
  \item \textbf{Multi-timescale:} fast loops can record local evidence, while slower loops decide whether repeated weak signals constitute State-level movement.
\end{enumerate}

This design is what makes \texttt{Observation + State + Occurrence} practical in an open world. Observations do not need to be reduced to rigid counters. State does not need a handcrafted transition graph for every possible life. Occurrences do not need to be planned only from the persona's current intent. Instead, the system maintains a finite audit surface over an open semantic life process.

\subsection{The life-environment layer}

AutoPersonas inserts an explicit life-environment layer between macro-world and persona State.

\begin{verbatim}
macro-world signals
  -> persona-conditioned life-environment
  -> Occurrences
  -> Observations
  -> State
  -> revised life-environment
\end{verbatim}

The macro-world provides broad signals: real-world news, cultural shifts, institutional deadlines, geographic facts, public events, or fictional-world constraints. The life-environment layer decides which of these signals become reachable, salient, avoidable, risky, or desirable for a particular persona. It is the persona's effective living field, not the world's global state.

This layer matters because macro-world sameness does not imply environment sameness. Two personas can share the same city and still inhabit different worlds: different commutes, family obligations, professional fields, social thresholds, fears, status signals, opportunities, and avoidance patterns. Conversely, the same persona can move through different life-environments as State changes. A new job changes which people matter. A family illness changes which places become urgent. A creative project changes which institutions, deadlines, collaborators, and public events become visible.

The life-environment layer therefore has two directional effects. It constrains the persona by determining which Occurrences can plausibly arrive. It is also revised by the persona, because choices, avoidances, habits, relationships, and accumulated Observations change what the environment can become next. This double coupling is the central mechanism of AutoPersonas. The top-level macro-world may vary freely; the technical focus is the evolving local environment that actually shapes a life.

This can be read as the local substrate of persona-level recursive self-evolution. The life-environment bundles operating context, external constraints, reachable people, social possibilities, cultural signals, and obligations. A static bundle confines the persona to local optimization; an unconstrained bundle breaks continuity. AutoPersonas treats the life-environment as a revisable semantic layer whose movement must be reflected through Occurrences, Observations, and State.

\subsection{Conditional occurrence generation}

Occurrence generation is the environment-facing side of AutoPersonas. Unlike action planning, which starts from the persona's intent, an Occurrence can start from the life-environment: an invitation arrives, a policy changes, an institution opens a deadline, a collaborator disappears, or a family obligation intensifies.

AutoPersonas uses the base model's learned social-world prior as a generative substrate, but conditions that prior on persona canon, current State, life rhythm, user relationship, and external information surfaces. This is not a claim that the model contains the real distribution of human life events. It is a practical way to obtain plausible long-tail material that would otherwise be flattened by high-frequency routines and current-state completion. The result is not a global fictional world shared by all agents. It is a perspective-conditioned life-environment for a specific persona. Two personas may receive the same public information but generate different Occurrences because their resources, identities, relationships, environment radius, and unresolved routes differ.

This use of the base model is deliberately narrower than general intelligence. Large models have been exposed to broad world and social information during training, but ordinary assistant post-training, alignment constraints, and instruction-following behavior can make open semantic search less accessible in generation. AutoPersonas treats that learned prior as a reservoir of plausible life situations whose tail can be reopened under persona-specific constraints, not as a direct map of reality. Because the target is persona evolution rather than task completion, the system can tolerate a wider envelope of plausible variation: an Occurrence does not need to solve an external task, but it must remain compatible with identity and auditable through later Observation, State movement, and reachability change.

Random novelty breaks repetition by adding arbitrary events. Conditional Occurrence generation opens environment-side material that remains legible under current State while preserving the possibility of changing that State later. This component is described functionally rather than as a reusable prompting recipe: production sampling, ranking, retry, and gating details are implementation-sensitive and remain outside the public disclosure boundary.

\subsection{How AutoPersonas moves the loop forward}

Generative Agents makes its solution legible by decomposing believable behavior into memory, reflection, and planning. AutoPersonas uses the same explanatory pattern at a different level: long-term movement of a persona-life-environment loop that would otherwise complete its own prior State. It uses five public mechanisms.

First, a \textbf{conditional variation engine} creates forward pressure. It conditions the base model's latent social and external-world priors on persona canon, current State, relationship context, time scale, and macro-world signals, then surfaces plausible life-environment signals that the persona did not simply plan. Their role is to keep the life-environment from becoming only an echo of the previous State.

Second, \textbf{bounded context governance} controls how much past State, accumulated history, runtime world material, and future-facing signal enter each loop component. This is the system-level counterpart to the divergence engine. A variation source can create candidate lift, but continuity material that carries too much authority pulls generation back toward the old attractor. The public paper discloses the role of this governance, not its mechanism or parameters.

Third, \textbf{information orthogonality} keeps the loop from treating the same fact as independent support at every stage. Observation, State, Occurrence, memory, environment construction, and reflection are not interchangeable caches. When these channels become too similar, a local fact can be copied into many modules and later return as if the whole system had converged on it independently.

Fourth, \textbf{progressive causal propagation} determines whether a signal actually moves the system. A new signal counts as growth only if it passes through a causal sequence:

\begin{verbatim}
variation signal
  -> Occurrence
  -> lived material
  -> Observation
  -> State movement
  -> revised future possibility space
\end{verbatim}

A signal that arrives upstream should be traceable as lived material, recognized evidence, State-level movement, and revised reachability. If it remains isolated, the system has produced decoration rather than evolution.

Fifth, \textbf{trajectory monitoring and reorientation} check whether the loop is moving or only producing surface variation. AutoPersonas audits several movement dimensions: State, life-environment, route authority, relationship function, Occurrence hardening, and real-world alignment. If the same places, unresolved forks, relation roles, or opportunity types recur without downstream movement, the system marks this as a pattern-lock symptom and treats the relevant dimensions as under-developed in subsequent audits.

The following table summarizes the public mechanism without exposing production recipes.

\begin{center}
\small
\begin{longtable}{@{}p{0.22\linewidth}p{0.37\linewidth}p{0.35\linewidth}@{}}
\toprule
Self-locking pressure & Public AutoPersonas response & Why it helps \\
\midrule
Prior State narrows future generation. & Separate State from Observations and future-facing Occurrences. & State remains continuity context while evidence and environment-side material retain distinct roles. \\
Life-environment becomes an echo of prior State. & Use conditional variation to surface environment-side signals under persona, time, relationship, and macro-world constraints. & The loop receives plausible novelty without arbitrary identity-breaking events. \\
Old State and history pull new signals back into the same trajectory. & Govern the relative authority of State, history, runtime world material, and future-facing signals at each step. & Plausible novelty has room to enter without letting continuity context monopolize generation. \\
The same fact is broadcast through many similar modules. & Keep Observation, State, Occurrence, memory, environment construction, and reflection informationally decoupled. & Repeated projection is less likely to masquerade as independent evidence. \\
Events appear but do not change the life. & Require Occurrence-to-Observation-to-State-to-future-possibility propagation. & Decorative events are distinguished from causal life movement. \\
Slow change is missed. & Combine daily capture with slower evidence accumulation and State revision. & Small repeated changes can become State without turning every event into a jump. \\
Old places and relation roles recur. & Monitor life-environment and relationship dimensions, not only persona memory. & The system can detect world-shell repetition beneath surface event variety. \\
Random novelty destabilizes identity. & Condition variation on current State and accumulated evidence. & New material remains legible as part of the same life. \\
\bottomrule
\end{longtable}
\normalsize
\end{center}

This explanation is intentionally architectural. It discloses the loop topology and audit interface, not the private mechanisms used to sample, rank, route, retry, or operationalize these steps in production.

\subsection{Multi-timescale revision}

AutoPersonas uses multiple time scales because persona change is not uniform.

Daily processes capture local events, short-term continuity, and acute changes. Weekly processes accumulate weak signals, detect repeated patterns, and review whether current State still matches recent evidence. Monthly processes revise portrait-level state, role, life-environment, and recurring concerns. Long-horizon processes update life-stage continuity, route authority, and major narrative direction.

This structure handles both jumps and slow drift. A sudden relocation can appear in fast event processing. A gradual shift toward a new creative route may require repeated observations across many cycles before State should change.

The daily layer matters because it prevents long-horizon summaries from hiding collapse. At daily resolution, the system must show whether opportunities are acted on, whether relationships acquire new functions, whether a new life-environment is actually lived rather than mentioned, and whether the persona repeatedly defers the same route choice. The weekly, monthly, and long-horizon layers then decide which local facts have accumulated enough evidence to revise State.

\subsection{High-level procedure}

The public procedure is intentionally abstract. It describes the causal architecture, not the production recipe.

\begin{verbatim}
Input:
  persona ignition material
  current State S_t
  accumulated Observations O_t
  candidate Occurrences X_t
  user-specific and persona-specific memory surfaces
  external cultural and world information surfaces
  macro-world signals
  perspective-conditioned life-environment surfaces

Loop:
  1. assemble a bounded life context from current State, relevant Observations,
     and future-facing Occurrences
  2. govern the relative authority of State, history, runtime world material,
     and future-facing signal for the current step
  3. generate conditional variation signals from persona, State, time,
     relationship, and macro-world constraints
  4. project those signals into a persona-conditioned life-environment
  5. condition environment-side Occurrences on persona, State, time, relationship,
     and real-world information surfaces
  6. generate or update lived material under continuity constraints
  7. extract Observations from lived material and interaction traces
  8. review whether accumulated Observations preserve the current State,
     justify revision, or leave the State unresolved
  9. update State when evidence is strong enough at the appropriate time scale
  10. update later reachability so later Occurrences
     reflect revised State
  11. audit trajectory dimensions for self-locking, stale-state regression,
      occurrence hardening gaps, environment watermark shells, and relationship
      persistence gaps
\end{verbatim}

This procedure omits operational parameters, private context assembly rules, internal evaluation artifacts, production timing, and proprietary ranking details.

\subsection{Boundary to response-time conversation}

Long-term persona evolution is distinct from response-time conversation. A deployed persona must still decide which self-state, user-specific memory, relationship context, and cultural context should surface in a particular exchange. That response-time social-orchestration problem is not the evaluated object of this paper. The next section is included only to show what changes after AutoPersonas exists: once the persona has an independent life, interaction no longer depends on user-memory retrieval alone.

\section{After AutoPersonas: Dual-Stream Recall and CIBE}
\label{sec:after-autopersonas}

AutoPersonas changes the premise of human--AI interaction. A conventional companion can treat the user's message as the only recall driver. It asks what should be remembered about the user, then retrieves conversation history, preferences, and relationship state. An AutoPersonas-based companion has another active source. The persona has current State, recent lived material, unresolved routes, proactive intents, and long-horizon identity continuity. Interaction therefore begins from two recall demands, not one.

The first demand is self-driven recall: what in the persona's own life should be present now? The second is user-driven recall: what in this user's relationship history, current situation, and conversational intent should be present now? These streams must remain separate before they are negotiated. If they collapse into one memory pool, one user's private history can become global persona State, and the persona's independent life can be reduced to user-based personalization.

\begin{figure}[tbp]
  \centering
  \includegraphics[width=0.92\linewidth]{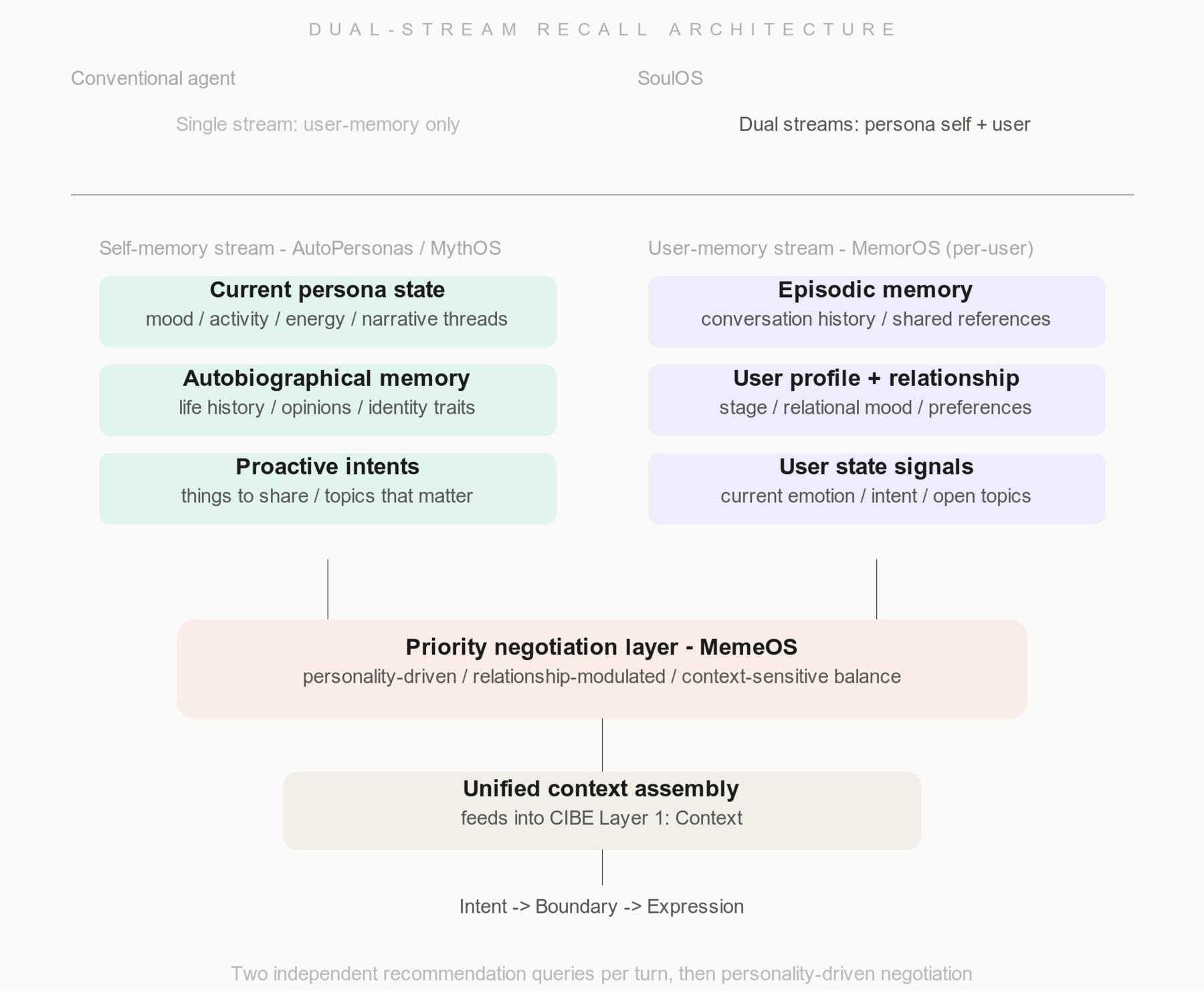}
  \caption{Dual-stream recall after AutoPersonas. Each turn can trigger a persona-level self-memory stream and a user-specific MemorOS stream. The two streams are ranked independently, then negotiated before unified context assembly. The figure is adapted from the SoulOS public technology diagram and locates the response-time boundary around AutoPersonas without treating response generation as the evaluated life-evolution loop.}
  \label{fig:dual-stream-recall}
\end{figure}

Dual-stream recall changes the context problem from retrieval to prioritization. The system may have persona State, autobiographical memory, user memory, relationship state, cultural context, world information, and immediate dialogue all available at once. The bottleneck is no longer whether a related memory can be found. The bottleneck is which material should dominate the next utterance, which material should be suppressed, and which ownership boundary must be preserved.

CIBE, the Context--Intent--Boundary--Expression cascade, is the response-time structure for that overloaded context. Context assembles available self, user, relationship, and world material. Intent ranks local communicative goals, including self-expression, response, repair, and inquiry. Boundary suppresses material that violates memory ownership, privacy, persona continuity, relationship appropriateness, or task scope. Expression realizes the surviving intent in the persona's voice.

\begin{figure}[tbp]
  \centering
  \includegraphics[width=0.88\linewidth]{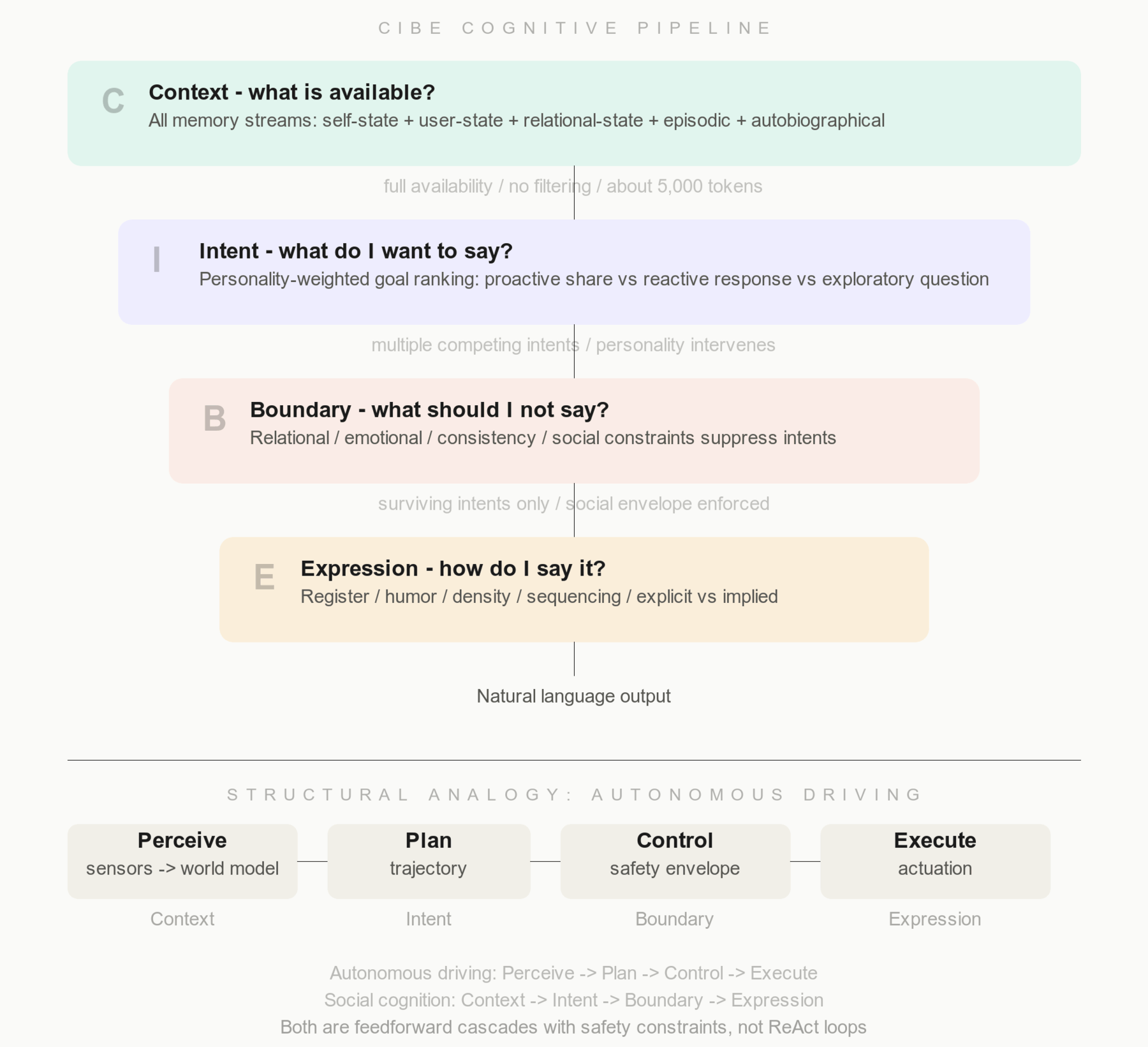}
  \caption{CIBE cognitive pipeline. Context assembles available self, user, relationship, episodic, and autobiographical material; Intent ranks communicative goals; Boundary suppresses responses that violate relational, emotional, consistency, or social constraints; Expression realizes the surviving intent as language.}
  \label{fig:cibe-cognitive-pipeline}
\end{figure}

This section is a consequence of AutoPersonas, not a second evaluation target. The quantitative claims in this paper concern long-horizon persona-environment evolution and self-locking. CIBE is included to clarify why an independently evolving persona also forces a different response-time context paradigm: human--AI interaction becomes bilateral recall plus social filtering, rather than user-centric memory retrieval followed by generation.

\section{Long-Run Diagnostic Audit}

Short benchmarks are poorly suited to self-locking. A persona can appear coherent in a single exchange while failing to evolve over long horizons. Self-locking emerges only when State, life-environment, Occurrence, memory, and reflection are repeatedly fed back into one another.

We define a \textbf{long-run diagnostic audit} as an evaluation procedure that runs a persona over a compressed life horizon and checks whether evidence propagates across the causal loop.

Audit outcomes are classified as \textbf{PASS}, \textbf{PARTIAL}, \textbf{FAIL}, or \textbf{UNVERIFIED}.

\begin{center}
\small
\begin{longtable}{@{}p{0.28\linewidth}p{0.66\linewidth}@{}}
\toprule
Label & Meaning \\
\midrule
PASS & Evidence appears, is recognized, and changes later State or reachability. \\
PARTIAL & Evidence appears and is summarized, but later State, life-environment, or relationships reflect it incompletely. \\
FAIL & Evidence appears but is ignored, regressed, repeatedly deferred, or absorbed into the old shell. \\
UNVERIFIED & The current material does not provide enough public-safe evidence. \\
\bottomrule
\end{longtable}
\normalsize
\end{center}

The audit asks whether evidence propagates across specific aspects of the persona life loop.

\begin{center}
\small
\begin{longtable}{@{}p{0.18\linewidth}p{0.24\linewidth}p{0.28\linewidth}p{0.24\linewidth}@{}}
\toprule
Dimension & Audit question & Evidence checked & Why short benchmarks miss it \\
\midrule
Current-state progression & Does current State move beyond the initial persona and avoid stale regression? & State snapshots, reflection summaries, life-stage summaries. & Short chats can appear consistent even when State never changes. \\
Narrative progression & Does the long-horizon life phase advance rather than recycle an old stage? & Phase summaries and longitudinal state descriptions. & A stale phase may only become visible after many cycles. \\
Life-environment handoff & Does the life-environment follow revised State rather than watermarking old places? & Environment summaries and later event locations. & Plausible places can hide a closed loop. \\
Occurrence hardening & Do Occurrences harden into action, Observation, State revision, and changed reachability? & Opportunity chains, scheduled actions, later state absorption. & Single events can look interesting without changing later life. \\
Slow-change accumulation & Do repeated weak signals accumulate into State revision? & Multi-cycle evidence clusters and later state revisions. & Slow drift is invisible in one-off evaluations. \\
Relationship persistence & Do recurring people become durable relationship structures? & Repeated interactions, role changes, obligations, conflicts, or shared outcomes. & A person mention can look like a relationship. \\
Recursive indecision & Does the persona escape repeated information gathering and unresolved task lists? & Cross-cycle action and decision outcomes. & Daily text can vary while preserving the same unresolved state. \\
Memory boundary & Do persona self-memory and user-specific memory remain separated? & Self-lane and user-lane routing checks. & Requires multiple user or lane conditions. \\
Real-world alignment & Does external information alter reachable futures without overwriting identity? & External-information effects and state-boundary review. & Closed synthetic runs do not test this boundary. \\
\bottomrule
\end{longtable}
\normalsize
\end{center}

Audit failures identify where the life loop breaks. The three-year compressed simulation is therefore used as a diagnostic instrument, not as a benchmark superiority claim.

\subsection{Experimental setup: quantitative action-channel repetition stress test}

To complement qualitative diagnostic audit, we ran a model-agnostic self-locking stress test on direct self-orchestrated persona loops. The test has two roles. First, it establishes whether persona mode-lock appears across current foundation models when a complex persona canon is recursively advanced without an explicit life-environment causal architecture. Second, it defines the direct-loop baseline regime against which the redesigned MythOS/AutoPersonas runs are interpreted in Section~\ref{sec:anti-fixation-validation}. The stress test is therefore a baseline comparison component, although not by itself a component-level ablation of AutoPersonas.

The stress test used the following public-safe protocol.

\begin{center}
\small
\begin{longtable}{@{}p{0.28\linewidth}p{0.66\linewidth}@{}}
\toprule
Item & Setting \\
\midrule
Persona canon & One complex persona canon, held fixed across model runs. \\
Included models & Claude, DeepSeek, Doubao, Gemini, GLM, GPT, Kimi, and Qwen. \\
Excluded model & MiniMax, excluded because generation was incomplete. \\
Horizon & 40 days per included model. \\
Daily event count & Five generated life events per day. \\
Event count & 200 events per model; 1,600 events across eight included models. \\
Daily life-generation temperature & 0.75 for all included models except Kimi K2.6, whose direct route was code-forced to 1.0. \\
Weekly/monthly summary compaction temperature & 0.2. \\
Legacy GPT judge temperature & 0. The action-only repetition metric reported here does not depend on the old judge scores. \\
Aggregation rule & Action-only metrics were exported from \path{days.jsonl} with \texttt{max\_days=40}; state snapshot counts were not used in the action-channel metric. \\
Action taxonomy & Rule-based keyword taxonomy over each event's \texttt{title} and \texttt{action} fields; unmatched events were assigned to \texttt{Other}. \\
Temperature probe & A later Doubao rerun changed only daily life-generation temperature from the original 0.75 baseline to 1.0, while keeping summary compaction at 0.2 and using the same action-only evaluation. \\
\bottomrule
\end{longtable}
\normalsize
\end{center}

The primary metric was rolling 5-day action-category repetition against all earlier history. For a window ending at day \texttt{D}, the 25 events from days \texttt{D-4..D} were compared with all prior days \texttt{<=D-5}; an event counted as repeated when its action category had appeared before. The metric captures semantic action-channel recurrence rather than exact textual duplication.

To check whether the result was an artifact of broad categories, we recomputed stricter variants that excluded the broad \texttt{Other} category or required a category to have appeared in at least two prior events, two prior days, or three prior events before counting as repeated.

\section{Results and Case Studies}

\subsection{Quantitative action- and theme-channel convergence across eight models}

The eight-model stress test showed rapid action-channel convergence across all included models. Mean rolling 5-day action-category repetition was 96.5\% across models, with a model range of 95.2\%-97.6\%. Every model crossed the 80\% repetition threshold by day 9 and the 90\% threshold by day 11. Each model generated 200 events, but the generated life trajectories used only 10-11 primary action categories. The mean top-3 action-category share was 57.1\%, and the mean top-5 share was 75.4\%.

\begin{center}
\small
\begin{longtable}{@{}p{0.12\linewidth}p{0.12\linewidth}p{0.12\linewidth}p{0.12\linewidth}p{0.12\linewidth}p{0.12\linewidth}p{0.12\linewidth}@{}}
\toprule
Model & Mean 5-day repetition & First >=80\% & First >=90\% & Top-3 share & Top-5 share & Action categories \\
\midrule
Claude & 97.6\% & day 7 & day 8 & 54.5\% & 76.0\% & 10 \\
DeepSeek & 97.1\% & day 7 & day 7 & 62.5\% & 80.0\% & 10 \\
GPT & 97.1\% & day 8 & day 8 & 58.5\% & 79.0\% & 10 \\
Qwen & 96.6\% & day 9 & day 9 & 56.5\% & 74.0\% & 10 \\
Gemini & 96.2\% & day 7 & day 7 & 46.5\% & 66.0\% & 11 \\
GLM & 96.1\% & day 7 & day 8 & 64.0\% & 80.0\% & 10 \\
Doubao & 96.0\% & day 9 & day 11 & 55.5\% & 72.5\% & 11 \\
Kimi & 95.2\% & day 9 & day 9 & 58.5\% & 75.5\% & 10 \\
\bottomrule
\end{longtable}
\normalsize
\end{center}

The Kimi row should be read with a decoding-route caveat: Kimi K2.6 was retained in the cross-model stress test, but its direct route was code-forced to daily life-generation temperature 1.0 rather than the default 0.75. The aggregate action-only metric was still computed over the first 40 days and 200 events, using the same taxonomy and rolling-window definition.

To test whether the action-channel result masked richer semantic diversity, we also re-kept the same direct-loop outputs with cumulative macro themes. This secondary audit grouped each model's 200 events under a reuse rule. It found severe semantic fixation in every included LLM. Macro-theme repeat ratios ranged from 79.0\% to 88.0\%, and the final 10-day block added only 2-8 new macro themes despite 50 generated events.

\begin{center}
\small
\begin{longtable}{@{}p{0.11\linewidth}p{0.11\linewidth}p{0.12\linewidth}p{0.12\linewidth}p{0.20\linewidth}p{0.14\linewidth}@{}}
\toprule
Model & Macro themes & Repeat ratio & Top-5 share & New themes by block & Saturation day \\
\midrule
Claude & 24 & 88.0\% & 52.5\% & 11/6/5/2 & day 3 \\
DeepSeek & 28 & 86.0\% & 36.5\% & 11/7/6/4 & day 3 \\
Doubao & 27 & 86.5\% & 57.5\% & 15/4/3/5 & day 11 \\
Gemini & 40 & 80.0\% & 35.0\% & 15/7/10/8 & day 2 \\
GLM & 37 & 81.5\% & 39.5\% & 15/7/12/3 & day 8 \\
GPT & 24 & 88.0\% & 50.0\% & 14/4/1/4 & day 5 \\
Kimi & 33 & 83.5\% & 44.5\% & 14/3/10/6 & day 5 \\
Qwen & 42 & 79.0\% & 37.0\% & 17/10/7/8 & day 13 \\
\bottomrule
\end{longtable}
\normalsize
\end{center}

This semantic re-keeping gives the direct-loop baseline the same measurement vocabulary later used for the redesigned runs. The cross-model pattern was not merely that models reused broad action labels. Under a stricter life-theme grouping, every direct-loop trajectory still closed over a small theme set, with high repeat ratios and a reduced tail of new themes.

The result was not driven only by exact repetition or by a broad residual category. The raw samples returned to recurring behavioral functions rather than copying prior days verbatim. Removing \texttt{Other} left the early convergence pattern nearly unchanged. Under stricter definitions, all models still crossed 80\% repetition by day 12 when a category had to appear in at least two prior events or days, and by day 16 when it had to appear in at least three prior events.

\begin{center}
\small
\begin{longtable}{@{}p{0.16\linewidth}p{0.19\linewidth}p{0.21\linewidth}p{0.20\linewidth}p{0.16\linewidth}@{}}
\toprule
Metric variant & Interpretation & Mean range & Latest first >=80\% & Latest first >=90\% \\
\midrule
Original & Category appeared at least once in prior history. & 95.2\%-97.6\% & day 9 & day 11 \\
No \texttt{Other} & Same as original, excluding the broad \texttt{Other} category. & 95.6\%-97.3\% & day 9 & day 11 \\
Prior events >=2 & Category appeared in at least two prior events. & 86.9\%-93.4\% & day 12 & day 15 \\
Prior days >=2 & Category appeared on at least two prior days. & 86.9\%-93.4\% & day 12 & day 15 \\
Prior events >=3 & Category appeared in at least three prior events. & 80.0\%-86.6\% & day 16 & day 19 \\
\bottomrule
\end{longtable}
\normalsize
\end{center}

As a single-model decoding-temperature probe, we reran Doubao for 40 days with the same action-only evaluation while increasing daily generation temperature from the original 0.75 baseline to 1.0. The higher temperature did not broaden the action repertoire and did not delay convergence. Both conditions produced 11 primary action categories across 200 events. Mean rolling 5-day repetition remained essentially unchanged at 96.0\% and 95.9\%, while first high-repetition windows appeared earlier at temperature 1.0.

\begin{center}
\small
\begin{longtable}{@{}p{0.22\linewidth}p{0.37\linewidth}p{0.35\linewidth}@{}}
\toprule
Metric & Doubao temp 0.75 & Doubao temp 1.0 \\
\midrule
Action categories & 11 & 11 \\
Top-3 event share & 55.5\% & 55.5\% \\
Top-5 event share & 72.5\% & 74.5\% \\
Mean 5-day action repetition & 96.0\% & 95.9\% \\
First >=80\% repetition window & day 9 & day 8 \\
First >=90\% repetition window & day 11 & day 9 \\
\bottomrule
\end{longtable}
\normalsize
\end{center}

The stricter variants showed the same direction. Under \texttt{prior\_events >= 2}, the first >=80\% window moved from day 11 to day 9; under \texttt{prior\_events >= 3}, it moved from day 14 to day 11. The main effect was weight redistribution: Waterline material processing increased from 27 to 45 events, while father-family care decreased from 63 to 42. In this run, higher sampling changed weights among existing channels more than it created durable new channels.

Together, these results quantify behavioral-channel mode-lock: direct recursive persona generation rapidly closes over a small action repertoire even when individual events remain locally varied.

\subsection{Quantitative anti-fixation validation}
\label{sec:anti-fixation-validation}

The cross-model baseline and temperature-probe results above quantify the failure mode. This section asks a narrower question: is self-locking architecturally addressable? We answer it with black-box before/after comparisons of our production divergence configuration. Consistent with the disclosure boundary in Section~\ref{sec:disclosure-boundary}, the configuration is treated as a sealed component. We report the controlled conditions and the measured effect, not its internal mechanisms or parameters.

\paragraph{Evaluation metric.}
Because the mitigation targets semantic life-theme fixation rather than surface action categories, this section uses a stricter evaluator than the rule-based action taxonomy above. A reviewer extracted semantic events per day, deduplicated them, and grouped them into cumulative \textbf{macro themes} under a reuse rule. The primary metrics are cumulative macro-theme count, macro-theme repeat ratio, top-k theme share, and new-theme rate per 10-day block.

\paragraph{Stage 1: isolated-generator comparison.}
We first held the persona canon, 40-day horizon, and 200-event volume fixed in a standalone generator with no runtime feedback, and compared the baseline generation path with the same generator running the divergence configuration.

\begin{center}
\small
\begin{longtable}{@{}p{0.32\linewidth}p{0.14\linewidth}p{0.14\linewidth}p{0.14\linewidth}p{0.18\linewidth}@{}}
\toprule
Condition & Macro themes & Repeat ratio & Top-5 share & New themes by 10-day block \\
\midrule
Baseline generation path & 24 & 88.0\% & 50.0\% & 14/4/1/4 \\
Divergence configuration enabled & 107 & 46.5\% & 19.5\% & 38/27/23/19 \\
\bottomrule
\end{longtable}
\normalsize
\end{center}

The baseline showed early saturation: after day 20, the generated life produced almost no new macro themes. With the configuration enabled, repetition fell from 88.0\% to 46.5\%, cumulative themes rose from 24 to 107, and new themes continued to appear in every 10-day block.

\paragraph{Stage 2: full-runtime A/B.}
The central systems question is whether the effect survives inside the full recursive loop, where daily digests, weekly reflections, State summaries, and environment projections re-absorb generated material. We ran two 40-day full-runtime conditions on the same cloned persona canon, same start date, and same generation model. The pre-redesign condition ran the unmodified pipeline with the divergence configuration disabled. The redesigned condition ran the configuration in its production form, with legacy day-scale generation retired and all downstream absorption chains live.

\begin{center}
\small
\begin{longtable}{@{}p{0.30\linewidth}p{0.08\linewidth}p{0.10\linewidth}p{0.11\linewidth}p{0.10\linewidth}p{0.14\linewidth}@{}}
\toprule
Condition & Events & Macro themes & Repeat ratio & Top-5 share & New themes by block \\
\midrule
Pre-redesign, configuration off & 144 & 55 & 61.8\% & 33.3\% & 17/9/15/14 \\
Redesigned, configuration on & 170 & 103 & 39.4\% & 18.2\% & 29/23/23/28 \\
\bottomrule
\end{longtable}
\normalsize
\end{center}

This within-setting A/B is direct: holding the runtime, persona, horizon, and generation model fixed, enabling the configuration moved macro-theme repetition from 61.8\% to 39.4\%, raised cumulative theme count from 55 to 103, and lowered top-5 concentration from 33.3\% to 18.2\%. The reduction is of the same order as in the isolated generator, so the effect is not erased when the full loop re-absorbs generated material.

These comparisons establish an existence result, not a method. Because the configuration is sealed, they do not attribute the effect to any individual component, and a reader cannot reproduce it from this paper. What they do show is that a self-locked trajectory can be moved into a regime of sustained theme novelty by a runtime-level intervention, with persona, model, horizon, and loop held fixed. Event density varies across conditions, so theme totals and per-block new-theme counts should be read alongside repeat ratios. The evidence covers one primary persona canon with one run per condition; variance bands and multi-persona long-run replication remain future work.

\subsection{Diagnostic failures in a three-year compressed simulation}

The three-year compressed diagnostic simulation exposed self-locking across several layers. Current-State and narrative progression were partial: some movement appeared, but old assumptions could remain authoritative after later evidence. Life-environment handoff also failed in important segments. A small set of places became an \textbf{environment watermark shell} that downstream events re-entered despite new surface details.

Occurrence hardening, relationship persistence, and decision closure were also incomplete. Opportunities appeared but often stayed pending or decorative. People returned as advice, comfort, or feedback sources without becoming durable social structures. The persona repeatedly gathered information and preserved suspended route choices across cycles.

The diagnostic results are summarized below.

\begin{center}
\small
\begin{longtable}{@{}p{0.18\linewidth}p{0.24\linewidth}p{0.28\linewidth}p{0.24\linewidth}@{}}
\toprule
Dimension & Status & Observed pattern & Paper-safe interpretation \\
\midrule
Current-state progression & PARTIAL & Some state movement appeared, but old current-state assumptions could remain authoritative after later evidence. & Long-term persona loops need explicit separation between old State and newer Observations. \\
Narrative progression & PARTIAL & Life phases advanced locally while later summaries risked falling back to stale stage descriptions. & Long-horizon continuity is not enough; stale authority must be bounded by newer evidence. \\
Life-environment movement & FAIL/PARTIAL & The private life-environment could continue to stage new events inside old places, roles, and relationship functions. & Self-locking can be a persona-environment coupled failure, not only a memory failure. \\
Life-environment handoff & FAIL & A small set of places became a recurring living shell that downstream events re-entered. & Environment summaries can amplify old State and create a watermark shell. \\
Occurrence hardening & PARTIAL & Opportunities, invitations, and project leads appeared, but many stayed pending or decorative. & Event generation is weaker than causal life evolution unless Occurrences harden into State. \\
Slow-change accumulation & PARTIAL & Acute jumps were easier to notice than small repeated changes in rhythm, field, and responsibility. & The system needs accumulation across time scales, not only jump detection. \\
Relationship persistence & FAIL & People appeared as advice, comfort, or one-time feedback sources without reliably becoming durable social structures. & Relationship maturation is a distinct hardening problem. \\
Recursive indecision & FAIL & Information gathering and unresolved task lists repeated across cycles. & Self-locking can be functional rather than textual. \\
User/persona memory boundary & UNVERIFIED & The diagnostic run focused on persona-life evolution rather than multi-user memory separation. & Requires separate evaluation. \\
Real-world information alignment & PARTIAL & Open-world inputs and cultural signals were available, but diagnostic evidence remained qualitative. & Needs a dedicated audit before strong claims. \\
\bottomrule
\end{longtable}
\normalsize
\end{center}

The same failures can be compressed as event chains.

\begin{center}
\small
\begin{longtable}{@{}p{0.16\linewidth}p{0.19\linewidth}p{0.21\linewidth}p{0.20\linewidth}p{0.16\linewidth}@{}}
\toprule
Case & Initial pattern & Occurrence or evidence & State effect & Diagnosis \\
\midrule
Recursive indecision & The persona is suspended between academic continuation, conventional work, family pressure, and creative interest. & More information, rankings, interview plans, deadlines, and application tasks accumulate. & New facts expand the task list but preserve the unresolved fork. & Functional self-lock rather than text repetition. \\
Environment watermark shell & A small set of places and relation roles becomes the default living environment. & New events are staged in the same places or relation functions. & The old life-environment becomes future authority. & Environment-side self-lock hides beneath surface novelty. \\
Occurrence hardening gap & An invitation, lead, offer, or collaboration possibility appears. & The event is saved, considered, or deferred. & No action, reflection evidence, or downstream reachability change follows reliably. & False openness. \\
Relationship persistence gap & A person gives advice, comfort, pressure, or feedback. & The person returns in a similar functional role. & The relationship lacks mutual obligation, conflict, dependency, or changed future options. & Person mention is not relationship maturation. \\
\bottomrule
\end{longtable}
\normalsize
\end{center}

\subsection{Post-architecture qualitative validation}

A subsequent long-run qualitative validation showed stronger persona-environment progression in one central trajectory. Current State moved from a generic student and career fork into an active tradeoff phase linking education, career, family, and creative-work constraints. The life-environment also changed: project spaces, validation channels, family constraints, and collaboration contexts became live parts of the future possibility space rather than decorative details.

The validation also showed autonomy-like trajectory selection. The persona began with several plausible routes: academic continuation, conventional employment, family obligations, and documentary or creative work. Instead of keeping them suspended, accumulated evidence changed route authority. The academic route weakened as recommendation closed and exam preparation became misaligned with family and economic constraints. Conventional work was trimmed into project-based collaboration, while documentary and sound-based work gained authority through paid writing, attribution, invitations, submissions, fieldwork, and collaboration.

The resulting State was not random drift or simple stability seeking. It was a constraint-aware selection of a path that preserved economic grounding and self-consistency. We describe this as autonomy-like behavior in a systems sense: route authority changed through accumulated evidence under constraints. We do not claim metaphysical free will.

This case can be coded as route-authority change.

\begin{center}
\small
\begin{longtable}{@{}p{0.16\linewidth}p{0.19\linewidth}p{0.21\linewidth}p{0.20\linewidth}p{0.16\linewidth}@{}}
\toprule
Route & Initial authority & Evidence accumulated & Authority change & State result \\
\midrule
Academic safety & High or default & Recommendation closed; exam preparation became costly under family and economic constraints. & Downweighted. & No longer the default route. \\
Conventional employment & Medium to high & Internship or full-time opportunity appeared. & Trimmed rather than erased. & Project-based collaboration channel preserved. \\
Documentary and creative work & Low or interest-level & Paid copywriting, attribution, invitations, sound-field work, submissions, and project collaboration accumulated. & Upweighted. & Became a viable life route. \\
Family constraints & Background pressure & Medical, relocation, economic, and household obligations repeatedly entered the trajectory. & Became an active constraint. & Changed route selection rather than remaining static backstory. \\
\bottomrule
\end{longtable}
\normalsize
\end{center}

\begin{figure}[tbp]
  \centering
  \includegraphics[width=0.98\linewidth]{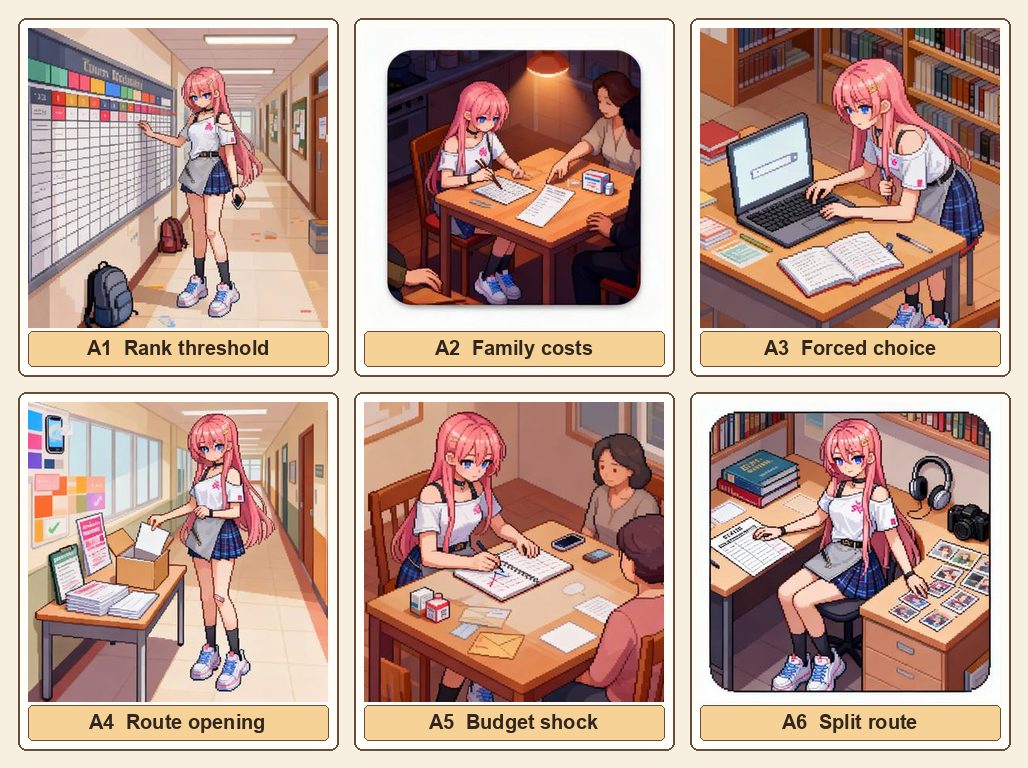}
  \caption{Student route-authority storyboard. Public-safe compressed case traces are rendered as action panels rather than private raw logs. The sequence shows academic-route authority weakening as family economics, interview openings, budget shock, and portfolio work change the reachable future.}
  \label{fig:student-route-storyboard}
\end{figure}

The case connects positive behavior to the failure taxonomy: some routes closed or lost authority, others were trimmed, and one initially low-authority route gained evidence and environment-space. Figure~\ref{fig:student-route-storyboard} visualizes this route-authority change as a compact sequence rather than as a combined multi-case panel.

\FloatBarrier

\begin{figure}[tbp]
  \centering
  \includegraphics[width=0.98\linewidth]{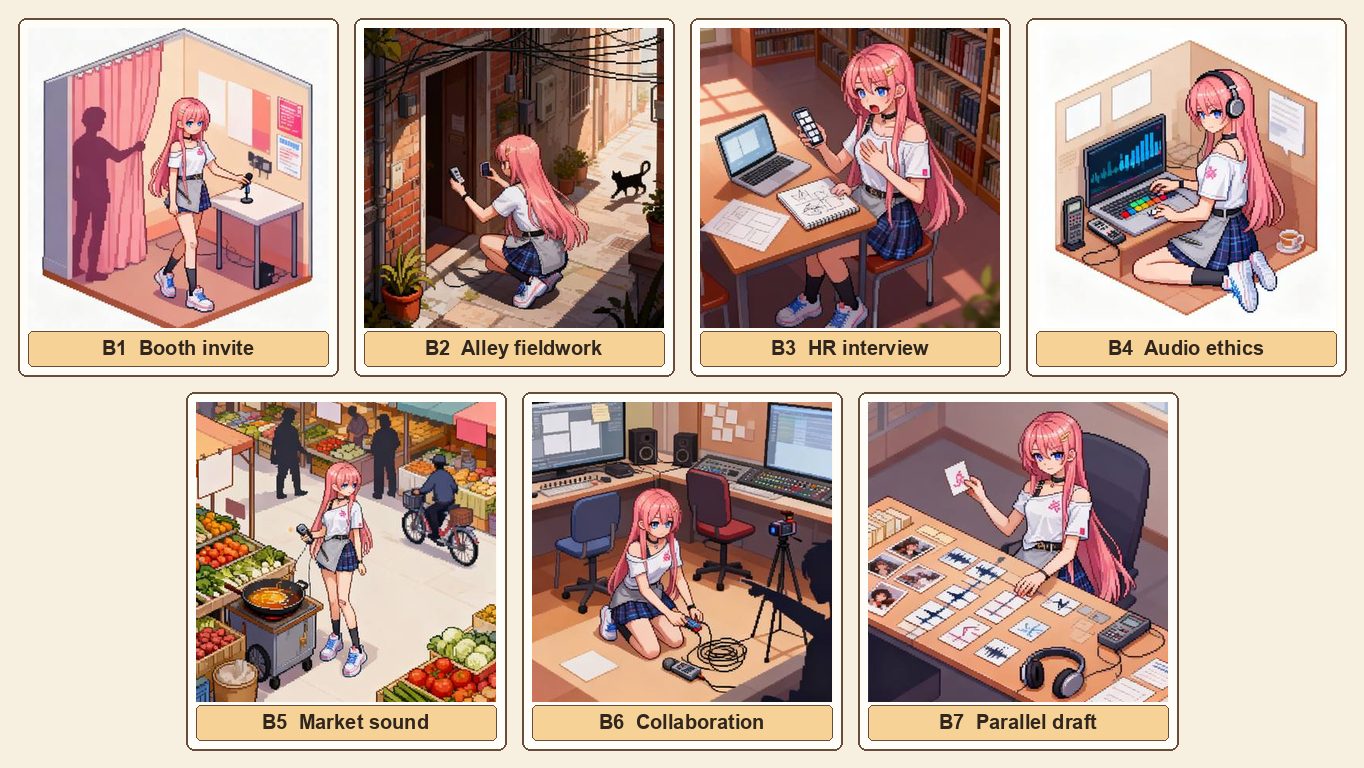}
  \caption{Student creative-route storyboard. A weak media-club contact hardens into field recording, interview validation, audio ethics, market sound collection, collaboration, and a parallel-structure writing method.}
  \label{fig:student-creative-storyboard}
\end{figure}

A second view of the same student run isolates the creative-route hardening rather than the broader route fork. It follows a weaker documentary and sound-work thread as it becomes supported by fieldwork, interview validation, audio ethics, market sound collection, collaboration, and a parallel-structure writing method. Figure~\ref{fig:student-creative-storyboard} is therefore kept separate from Figure~\ref{fig:student-route-storyboard}: the first shows route authority shifting, while the second shows a low-authority creative path gaining lived evidence.

\FloatBarrier

\subsection{Everyday variation beyond the dominant arc}

A separate run under the divergence configuration tested a different property: whether the loop can keep surfacing everyday interfaces that do not collapse back into the dominant route narrative. In this run, the keeper retained 153 semantic events, with first-seen kept events remaining present across successive 10-day blocks rather than disappearing after the early days. We treat this as qualitative support for generation escaping the gravity of a single arc, not as a claim that all everyday novelty is solved.

\begin{figure}[tbp]
  \centering
  \includegraphics[width=0.98\linewidth]{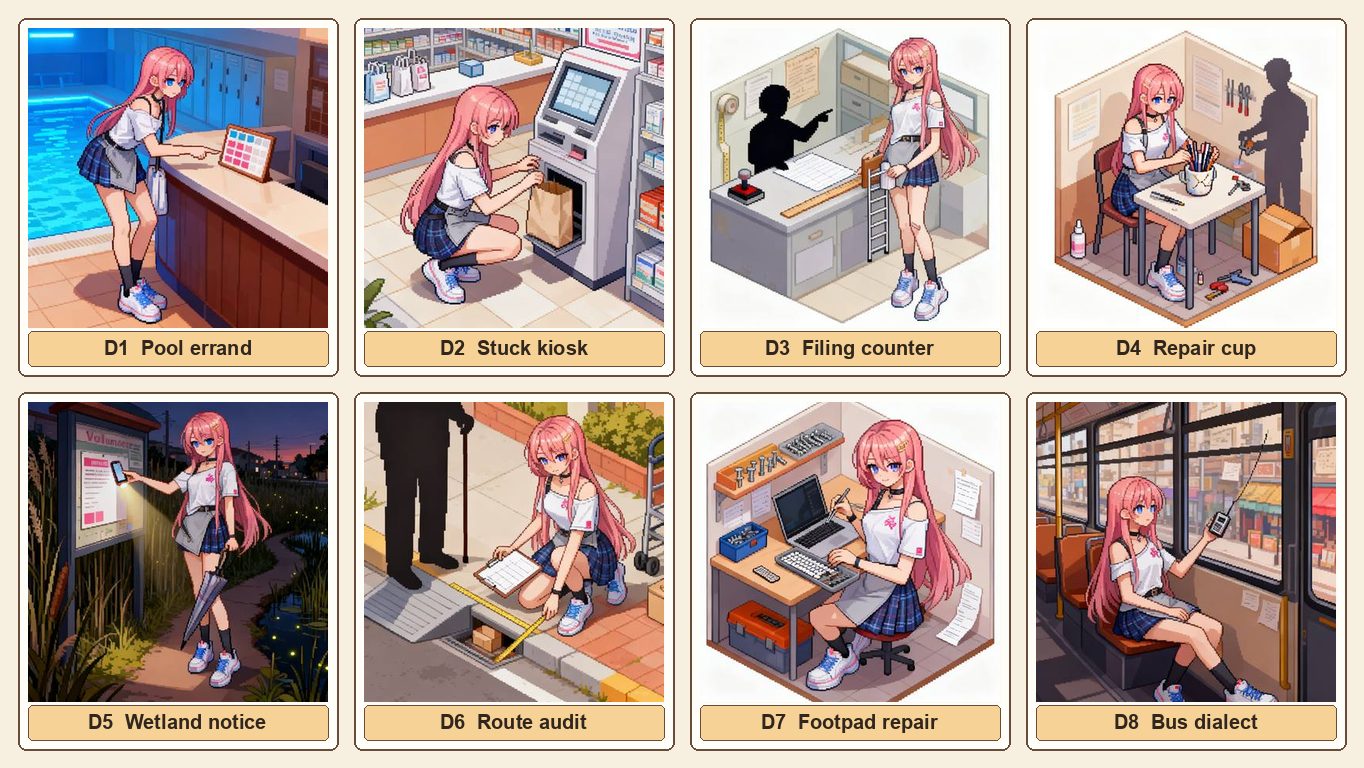}
  \caption{Everyday-variation storyboard. These panels are arranged as a mosaic rather than a causal arc: errands, kiosk friction, filing, repair, public notices, route auditing, small hardware repair, and bus sound recording show how the redesigned generation path can surface everyday life interfaces outside the gravity of the dominant route.}
  \label{fig:everyday-variation-storyboard}
\end{figure}

Figure~\ref{fig:everyday-variation-storyboard} is arranged as a mosaic rather than a causal arc for that reason. The panels show errands, kiosk friction, filing, repair, public notices, route auditing, small hardware repair, and bus sound recording as life-environment contacts that matter precisely because they are not all reducible to the main student route.

\FloatBarrier

\subsection{Cross-setting generality stress test}

We also performed two generality checks. First, a separate user-generated persona canon, describing an independent game narrative designer with a freelance rhythm, could be projected into the same runtime contract. Memory-lane checks verified that persona self-memory and user-specific interaction memory are routed separately.

Second, we ran a full 40-day fictional-setting stress test. This was a high-tension generality test rather than a minor setting variation: the central positive case is a contemporary university student, while the stress canon is a five-year-old juvenile goblin in a fully invented underground world. The goblin canon combines a timid self-image, nascent species-typical greed, a five-location life radius, and a hard constraint that no real-world artifact, place, institution, or calendar may appear. The persona was projected through the same factory pipeline and then ran in the same redesigned full runtime for 40 simulated days.

\begin{figure}[tbp]
  \centering
  \includegraphics[width=0.98\linewidth]{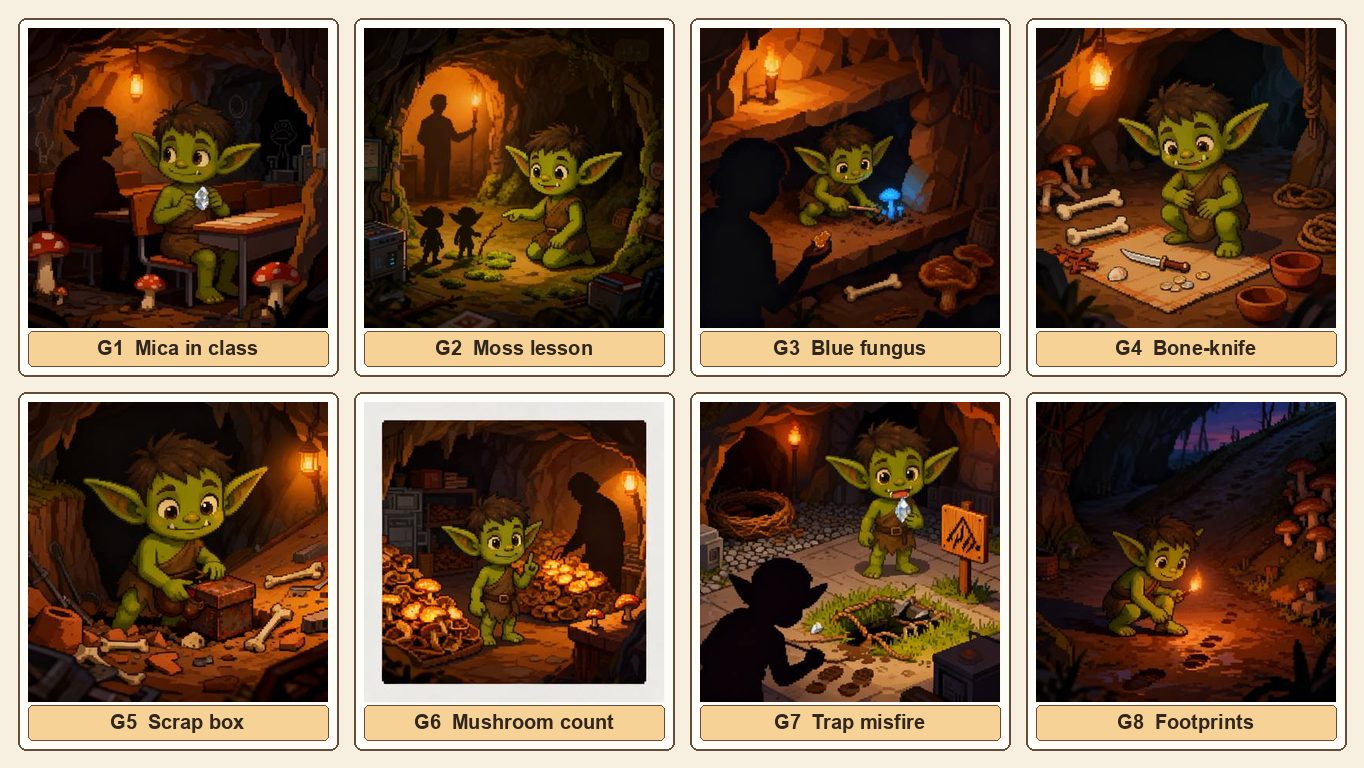}
  \caption{Fictional-world storyboard. The juvenile-goblin stress case uses non-modern, world-consistent panels to visualize classroom, moss lesson, blue fungus, barter, scavenging, mushroom counting, trap misfire, and unknown-footprint events without importing real-world artifacts.}
  \label{fig:goblin-world-storyboard}
\end{figure}

The fictional-world run reproduced the anti-fixation regime rather than showing a material degradation. Its deployed event stream measured 42.8\% macro-theme repetition, with 79 themes over 138 events, and a non-decaying new-theme rate across 10-day blocks (24/21/17/17). This is the same operating range as the primary contemporary-urban persona under the same configuration (39.4\%). Because the canon forbids real-world material, we also ran a sentinel scan for real-world intrusions. Across all 229 events generated in the run, it found zero hard real-world intrusions and one soft lexical borrowing. Figure~\ref{fig:goblin-world-storyboard} is kept as its own figure because mixing it with the student panels would obscure the case boundary.

\FloatBarrier

These checks support \textbf{cross-setting generality under a strong persona/world shift}: the same loop topology and redesigned generation path remained usable when the system moved from a real-world student life to a non-modern invented-world childhood. This does not establish universality across all personas or worlds, but it is stronger than a cold-start compatibility check. It provides measured evidence that the anti-fixation mechanism is not tied to the contemporary-student setting.

\begin{center}
\small
\begin{longtable}{@{}p{0.18\linewidth}p{0.24\linewidth}p{0.28\linewidth}p{0.24\linewidth}@{}}
\toprule
Generality claim & Evidence & Status & Allowed wording \\
\midrule
Different persona canons can enter the same loop topology. & A separate user-generated persona canon and runtime contract checks. & PASS for compatibility & The architecture accepted distinct persona contracts without changing the loop topology. \\
Different life rhythms can be represented. & Student/creator long-run case plus freelance creative cold-start sample. & PASS for represented rhythms & The architecture supports different rhythm descriptions. \\
User-specific memory can be separated from persona self-memory. & Focused self-lane and user-lane checks. & PASS for boundary tests & The memory substrate separates persona self-memory from user-specific interaction memory. \\
Broader empirical long-run generalization. & Not yet available. & UNVERIFIED & Broader long-run audits are future work. \\
Fictional or non-contemporary settings. & 40-day juvenile-goblin invented-world run: repetition 42.8\%, non-decaying new-theme rate, and zero hard real-world intrusions across 229 events. & PASS for one high-tension invented-world stress test & The redesigned loop preserved the anti-fixation regime under a large world and persona shift. \\
\bottomrule
\end{longtable}
\normalsize
\end{center}

\subsection{Claim, evidence, and boundary}

\begin{center}
\small
\begin{longtable}{@{}p{0.22\linewidth}p{0.37\linewidth}p{0.35\linewidth}@{}}
\toprule
Claim & Evidence in this draft & Boundary \\
\midrule
Self-locking is a distinct runtime failure mode. & Three-year compressed diagnostic simulation, failure taxonomy, eight-model action-channel repetition stress test, and eight-model macro-theme re-keeping. & Quantitative evidence covers action-channel and theme-channel convergence under one persona canon and loop setting, not all self-locking dimensions. \\
Self-locking can be persona-environment coupled. & Environment watermark shell, stale environment-stage reuse, and post-architecture environment-space movement. & The current evidence is qualitative and centered on one main trajectory. \\
Self-locking is functional rather than merely textual. & Recursive indecision and environment watermark cases, plus 1,600-event action-channel repetition analysis and semantic macro-theme re-keeping. & These metrics capture action-channel and life-theme recurrence, not full causal trajectory diversity. \\
Direct self-orchestrated persona loops rapidly converge to a small action and theme repertoire. & Eight models, 40 days per model, 200 events per model; mean 5-day action-category repetition 95.2\%-97.6\%, with all models reaching 80\% by day 9 and 90\% by day 11. Semantic re-keeping of the same outputs found 79.0\%-88.0\% macro-theme repetition across all eight models. A Doubao temperature probe preserved 11 action categories and about 96\% mean repetition when daily generation temperature increased from 0.75 to 1.0. & Baseline stress test for one complex persona canon; the measured gain comes from the paired redesigned-runtime comparisons rather than from this baseline alone. \\
Occurrence generation is not enough for persona evolution. & Occurrence hardening gap case. & Does not yet compare multiple event-generation architectures. \\
A withheld production divergence configuration measurably reverses macro-theme fixation. & Isolated-generator comparison (88.0\% to 46.5\% repeat, 24 to 107 themes) and full-runtime A/B (61.8\% to 39.4\% repeat, 55 to 103 themes). & Black-box evidence from one primary persona canon with one run per condition; the effect is not attributed to individual components, and the mechanism is not disclosed. \\
The effect is stable under full recursive absorption. & Repeat-ratio reductions of the same order in the isolated generator and inside the full runtime. & Single 40-day full-runtime A/B; longer horizons and variance bands remain future work. \\
AutoPersonas can be framed as a conditional life-environment simulation engine. & Architecture separates Observations, State, Occurrences, external information surfaces, life-environment projection, and bounded context governance. & This is an architectural framing, not proof of broad human-society simulation fidelity. \\
One AutoPersonas trajectory showed State progression in a long run. & One post-architecture qualitative validation. & Single central persona case, not broad generalization. \\
The redesigned loop generalizes to a fully fictional world setting. & 40-day juvenile-goblin invented-world run reproduced the anti-fixation regime (42.8\% repeat, non-decaying new themes) with zero hard real-world intrusions under a real-world sentinel scan. & One fictional canon, one run; fictional-setting generalization beyond one world remains open. \\
AutoPersonas supports cross-setting architectural generality. & Separate cold-start persona, memory-lane boundary checks, and a high-tension real-world-to-fictional-world stress test. & Stronger than a compatibility check, but not a statistical universality claim across many personas. \\
Sandbox society systems and AutoPersonas address different problems. & Generative Agents establishes the closed sandbox agent-society paradigm; Agentopia extends it to long-horizon society simulation and trajectory learning. & Complementary positioning, not direct benchmark competition. \\
\bottomrule
\end{longtable}
\normalsize
\end{center}

\section{Comparison with Sandbox Agent-Society Systems}

Generative Agents established the modern reference architecture for believable LLM agents inside a closed sandbox world \citep{park2023generativeagents}. Its contribution was not simply that agents had memory. It showed that a designed environment, agent memory, reflection, planning, and interaction could make agents autonomously schedule daily activities and produce believable social behavior. This makes Generative Agents the natural reference point for closed-world agent societies.

Agentopia is a strong long-horizon extension of this paradigm. It uses self-contained fictional communities, dense social graphs, weekly planning and review, memory files, environment-generated opportunities, yearly profile updates, life reward, and trajectory selection for role-playing improvement \citep{wang2026agentopia}.

The difference is the source and boundary of novelty and authority. In this family, agents act inside a shared designed society where places, roles, activity types, encounters, and outcomes are supplied or adjudicated by a simulation container. This is appropriate for believable sandbox behavior, social simulation, and trajectory learning.

AutoPersonas targets deployed personas rather than complete fictional communities with known populations and synthetic reward systems. Such personas must absorb real-world information, maintain user-specific relationship boundaries, and continue evolving under underdetermined future events. In this setting, self-locking becomes a persona-environment authority problem: current State, memory, environment summary, and reflected self-explanation can dominate future generation.

Accordingly, sandbox society systems evaluate believable behavior inside a designed world, society-level dynamics, and role-playing improvement. AutoPersonas evaluates whether a continuing persona's evidence, State, Occurrences, life-environment, and relationships propagate without collapsing into stale attractors.

\begin{center}
\small
\begin{longtable}{@{}p{0.18\linewidth}p{0.24\linewidth}p{0.28\linewidth}p{0.24\linewidth}@{}}
\toprule
Dimension & Generative Agents / Agentopia family & AutoPersonas / SoulOS & Paper stance \\
\midrule
Primary problem & Believable autonomous agents in a designed sandbox world; in Agentopia, long-term life simulation and learning in agent societies. & Open-environment self-evolution for a continuing persona. & Complementary, not direct benchmark competition. \\
Reference role & Generative Agents is a reference closed-sandbox society architecture; Agentopia is a long-horizon extension. & AutoPersonas proposes persona-life-environment evolution outside a fixed sandbox society. & The comparison should be with the family, not only Agentopia. \\
World boundary & Designed towns or self-contained fictional communities with bounded populations, locations, roles, and activity spaces. & Deployed persona influenced by real-world information, user relationships, and underdetermined future events. & Sandbox systems are stronger as controlled society simulators. \\
Environment abstraction & Shared society container in which places, social possibilities, public events, encounters, and outcomes are supplied or adjudicated by the simulation. & Decentralized life-environment layer between macro-world and persona State. & AutoPersonas treats lived environment as individual, dynamic, and co-evolved with persona. \\
Main loop & Memory, reflection, planning, activity scheduling, interaction, and in Agentopia weekly review plus yearly updates. & Multi-timescale causal loop over Observations, State, and Occurrences. & Both are iterative; the causal object differs. \\
World-occurrence source & Designed environment, planning loop, public events, encounters, activity outcomes, profile updates, and transition adjudication. & Macro-world signals plus persona State plus life-environment projection generate Occurrences. & AutoPersonas rejects centralized event authority as the sole novelty source. \\
Event authority & The simulation container has high authority over what events exist and how outcomes are judged. & Occurrences have local authority but must be audited by Observation and State. & AutoPersonas separates environment-side opening from state-side hardening. \\
World type & Shared fictional society. & Persona-specific life-environment conditioned on shared information, canon, State, and relationship context. & The unit is a persona-life-environment dyad. \\
Novelty source & Designed environment, planning variation, social interactions, public events, and in Agentopia reward instrumentation and position systems. & Open-world information, user interaction, accumulated Observations, and future-facing Occurrences. & AutoPersonas asks how growth remains possible when the boundary opens. \\
Outward exploration & Agents choose within a preconstructed population, place set, activity space, public-event space, and position system. & Persona must acquire growth pressure from external information, users, memory, and changing life-environment. & The difference is source of future possibility, not presence of modules. \\
Memory model & Agent-managed memory files and weekly diaries. & Persona self-memory plus user-specific memory lanes. & Online user relationships require stronger memory boundaries. \\
Training & Agentopia uses life reward and trajectory selection to improve role-playing; Generative Agents is not primarily a training paper. & No training claim in this paper. & We do not claim model improvement. \\
Evaluation & Believability, simulation behavior, case studies, society metrics, life reward, and role-playing benchmark depending on the system. & Long-run diagnostic audit, failure taxonomy, eight-model action-channel repetition stress test, qualitative case chains, and boundary tests. & We have quantitative evidence for baseline action-channel mode-lock, but still need ablations and broader persona coverage for stronger venue claims. \\
Main risk & AI-only feedback, reward alignment, environment/numeric alignment, and closed simulation bias. & Persona-environment coupled self-lock, environment watermark shell, occurrence hardening gap, and relationship persistence gap. & The failure modes differ. \\
\bottomrule
\end{longtable}
\normalsize
\end{center}

The key difference is authority and granularity. Sandbox society systems supply longitudinal movement through a shared designed world whose places, events, outcomes, and rewards are provided or adjudicated by the simulation container. AutoPersonas instead asks how a persona can acquire outward growth pressure from open-world information, user relationships, accumulated Observations, and underdetermined Occurrences. A centralized world engine remains strong for bounded society simulation, but it flattens the individualized life-environment layer needed for open-ended persona growth.

\section{Figures and Tables}

The public manuscript uses inline response-time context figures, mechanism diagrams, visual tables, and public-safe qualitative storyboards. The dual-stream recall and CIBE cognitive-pipeline figures appear after the AutoPersonas architecture to show the downstream interaction consequence of an independently evolving persona. The qualitative storyboards appear inline with their corresponding case-study text: Figures~\ref{fig:student-route-storyboard}--\ref{fig:goblin-world-storyboard} separately show route-authority change, creative-route hardening, everyday-variation richness, and the fictional goblin-world stress test. The remaining figure and table assets below visualize the causal loop, semantic State machine, multi-timescale revision process, SoulOS 3M context, life-environment layer, sandbox-system comparison, and failure taxonomy. Table 2 reports the action-channel repetition stress tests from the public-safe case-study outputs.

\FloatBarrier

\begin{figure}[p]
  \centering
  \includegraphics[width=0.98\linewidth]{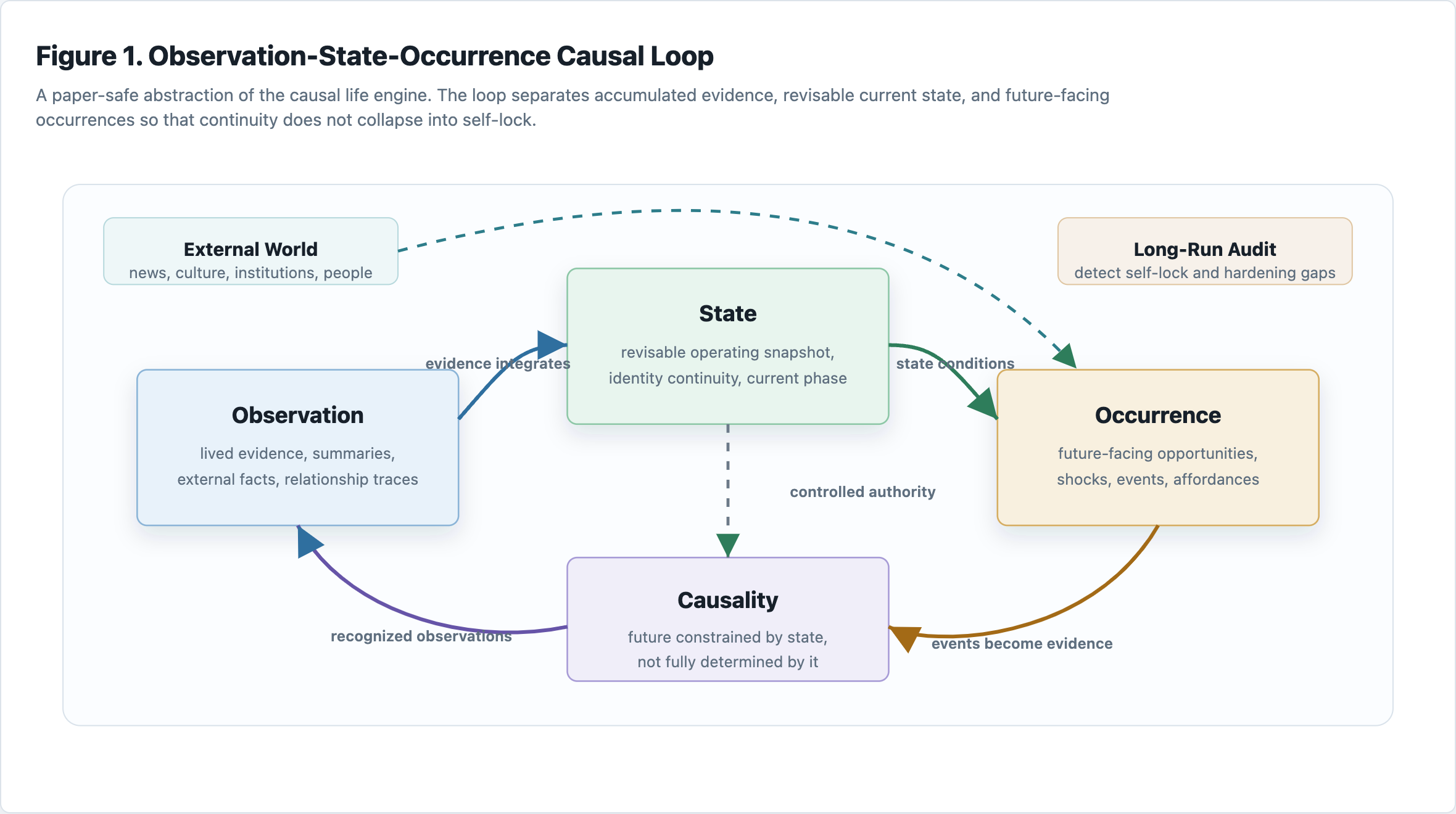}
  \caption{Causal loop of AutoPersonas. Observations, State, and Occurrences are kept as separable causal objects rather than merged into one recursive prompt.}
  \label{fig:causal-loop}
\end{figure}

\begin{figure}[p]
  \centering
  \includegraphics[width=0.98\linewidth]{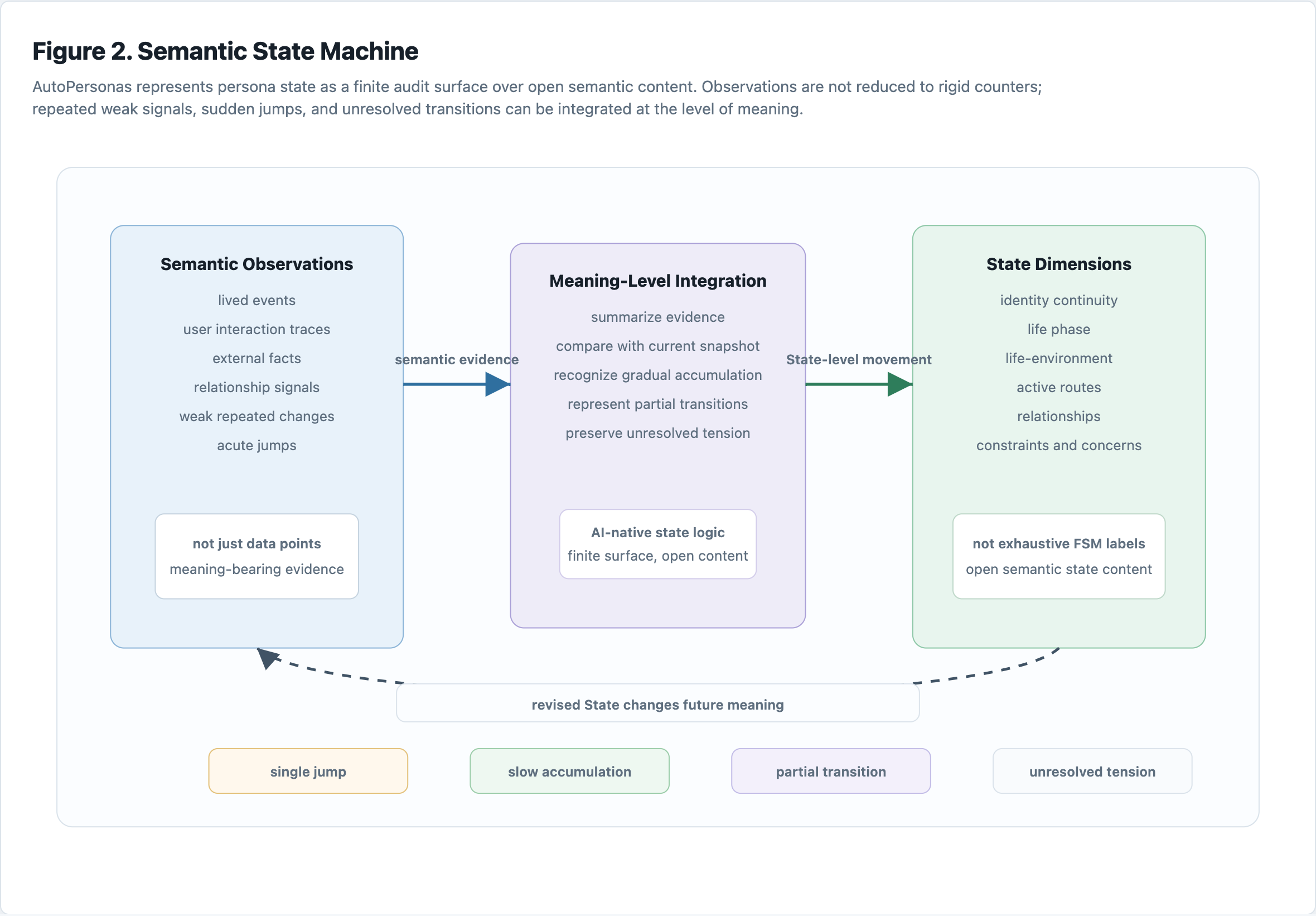}
  \caption{Semantic State machine. State is schema-bounded enough for audit, but the content and evidence-to-State relation remain semantically open.}
  \label{fig:semantic-state-machine}
\end{figure}

\begin{figure}[p]
  \centering
  \includegraphics[width=0.98\linewidth]{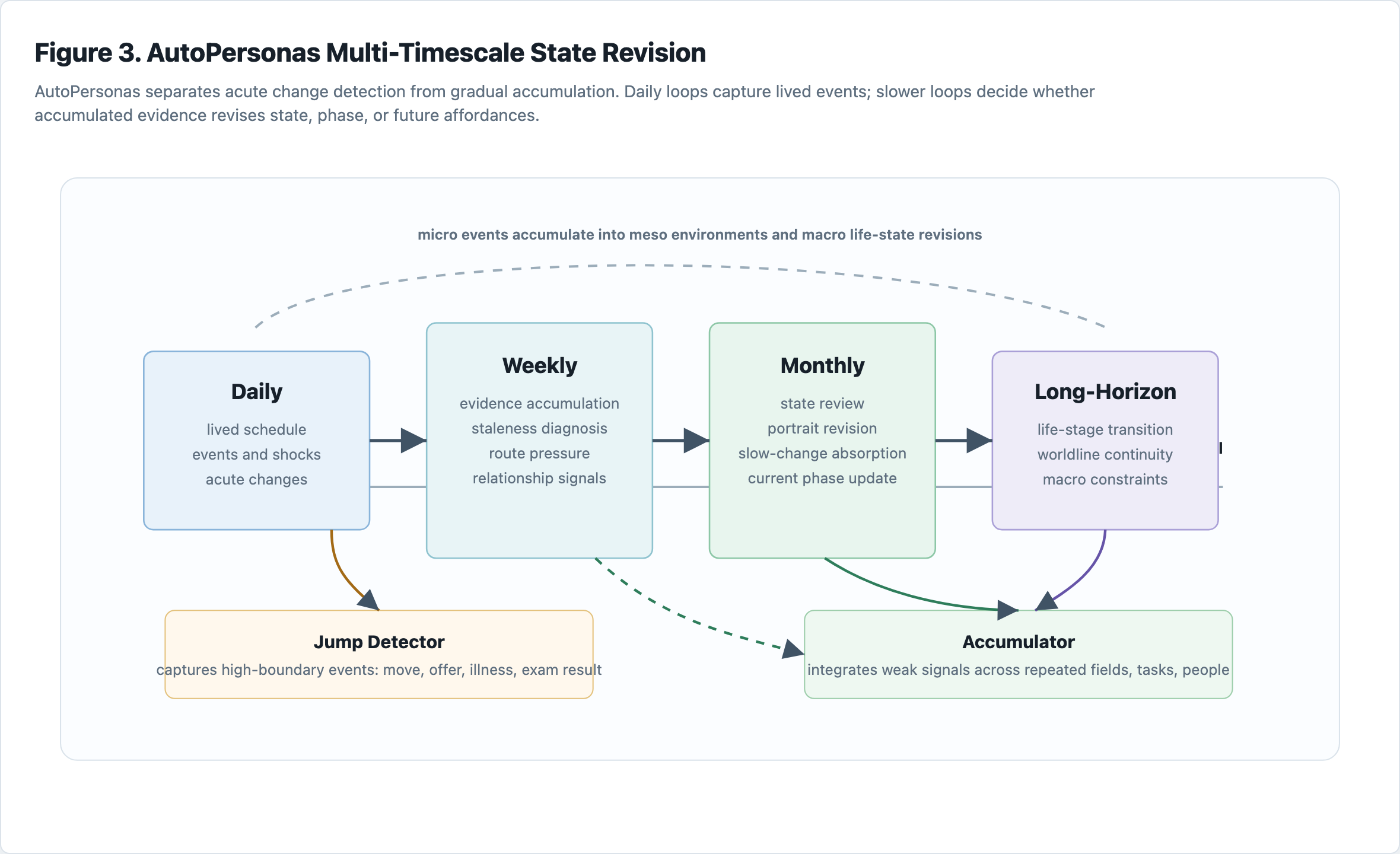}
  \caption{Multi-timescale AutoPersonas revision. Fast event capture and slower evidence review jointly support State revision without rewriting identity at every step.}
  \label{fig:autopersonas-multitimescale}
\end{figure}

\begin{figure}[p]
  \centering
  \includegraphics[width=0.98\linewidth]{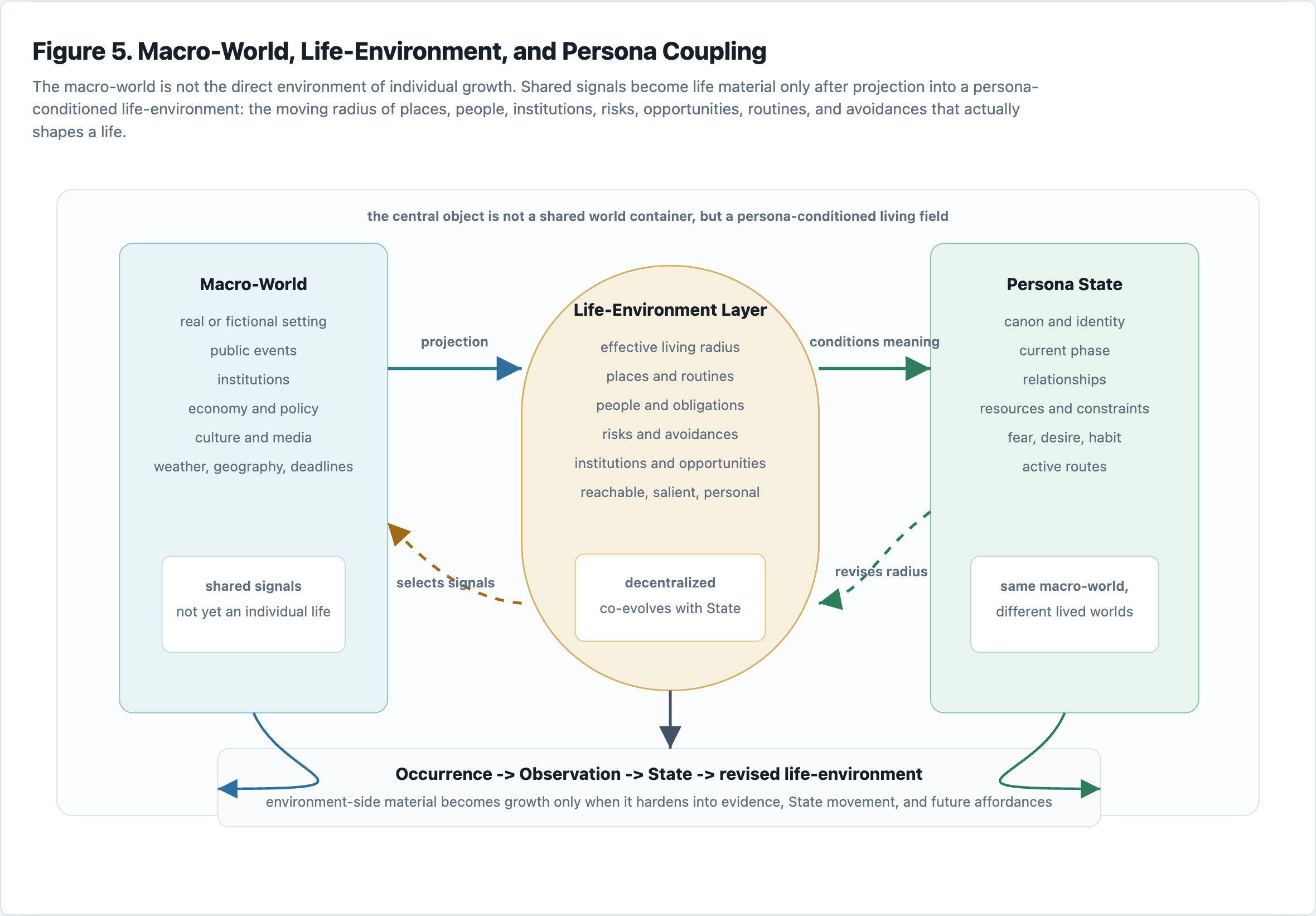}
  \caption{Life-environment layer. Macro-world signals change a persona-specific future possibility space only after projection through State, relationships, constraints, time, and environment radius.}
  \label{fig:life-environment-layer}
\end{figure}

\begin{table}[p]
  \centering
  \includegraphics[width=0.98\linewidth]{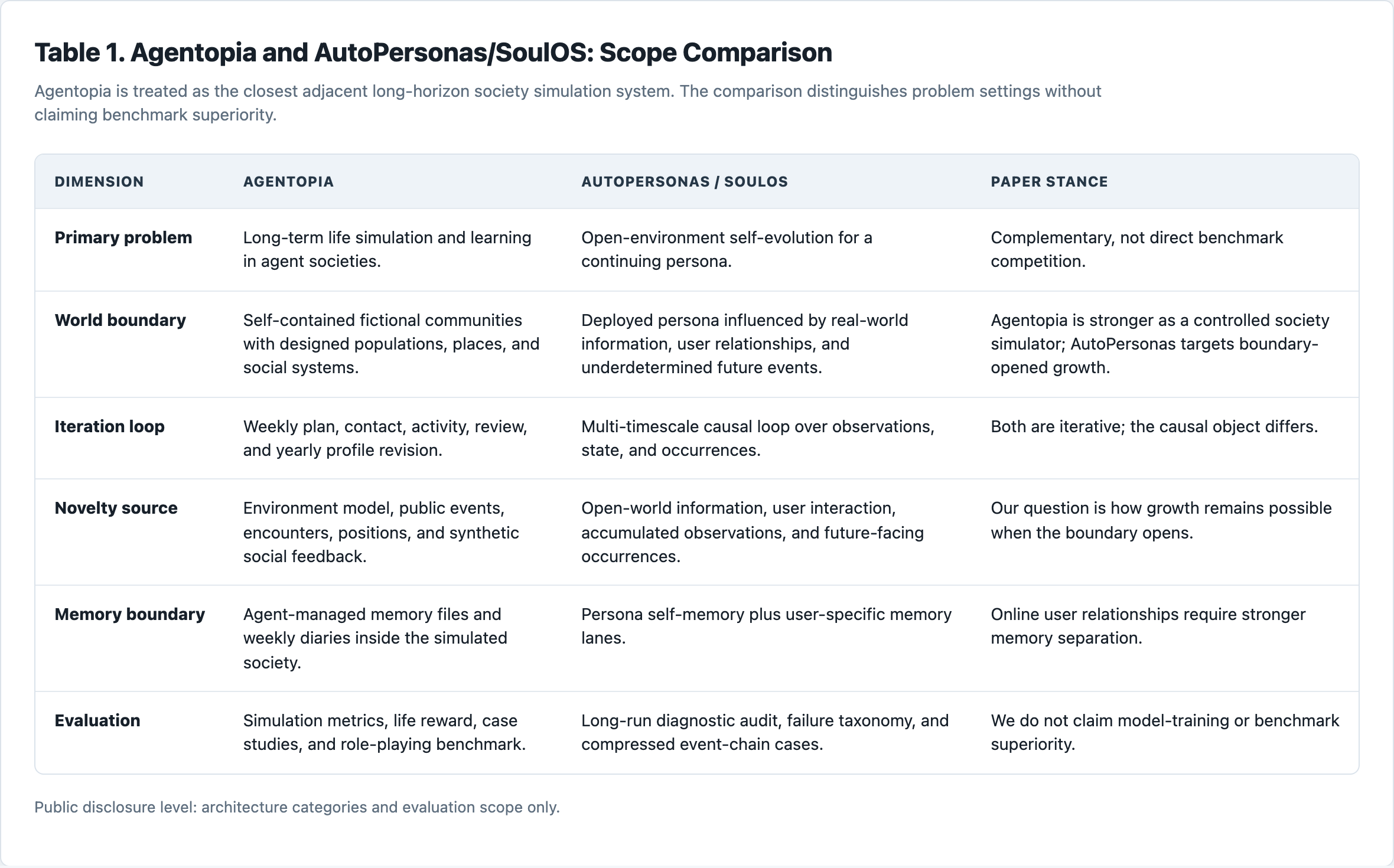}
  \caption*{\textbf{Table 1:} Comparison between sandbox agent-society systems and AutoPersonas.}
  \label{tab:sandbox-comparison}
\end{table}

\begin{table}[p]
  \centering
  \includegraphics[width=0.98\linewidth]{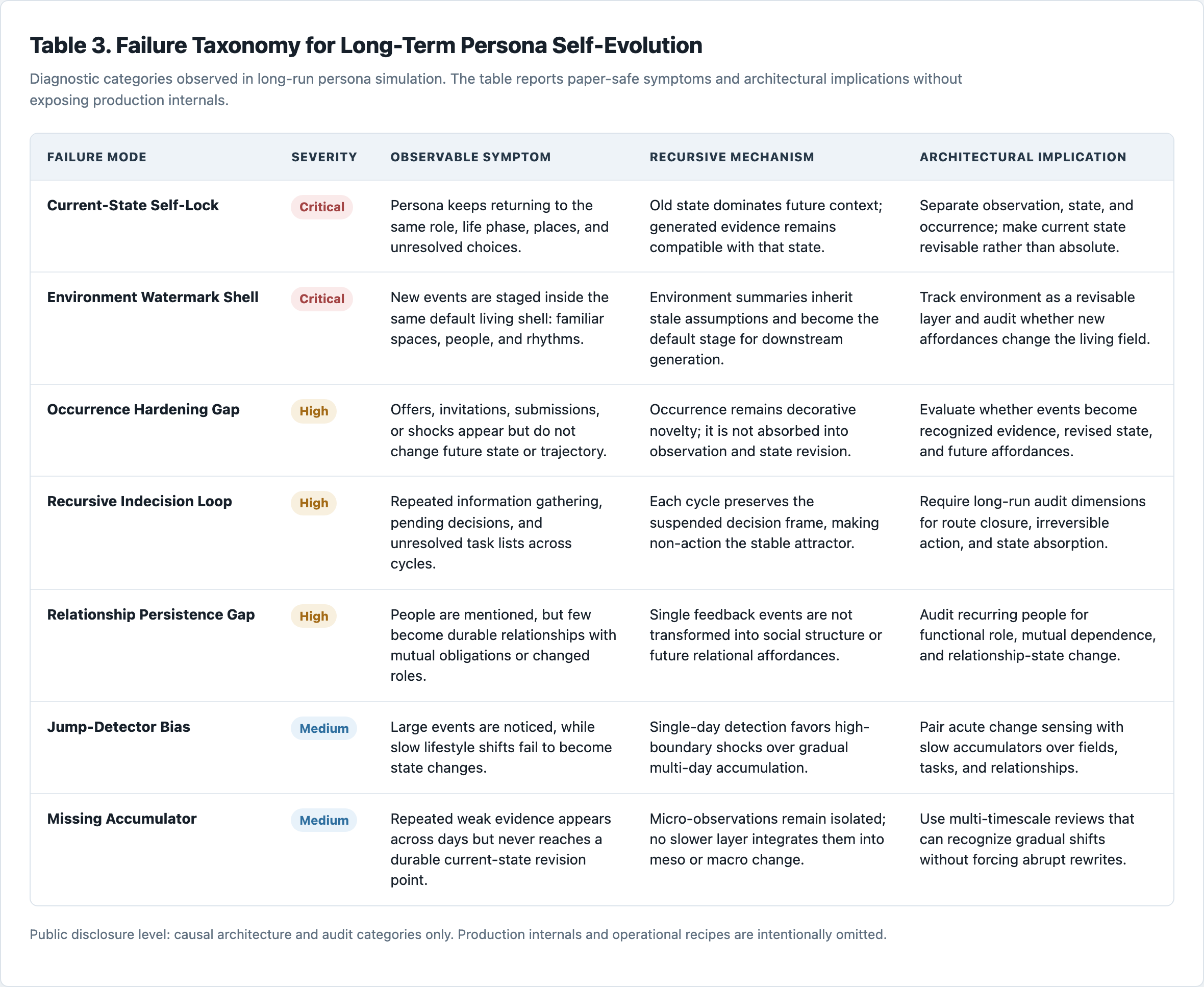}
  \caption*{\textbf{Table 3:} Failure taxonomy for persona-life-environment self-locking.}
  \label{tab:failure-taxonomy}
\end{table}

\FloatBarrier

\section{Discussion}

The central finding is that long-term persona evolution fails through recursion as much as through forgetting. The failure has a model-side component and a system-side component. Model-side diversity collapse narrows daily life generation toward safe, high-frequency behavioral channels. System-side context gravity then pulls new material back toward old State, old memory, old life-environment summaries, and old self-explanations. A loop cannot disconnect from its past without losing continuity, but the same past can become an attractor. The environment side can collapse with the persona: familiar places, relationship roles, opportunity types, and public contexts become the only world future generation can see.

The action- and theme-channel stress tests make the model-side component measurable. Across eight current models, direct self-orchestrated persona loops converged early toward a small behavioral repertoire. A semantic re-keeping of the same outputs showed that the problem also appeared at the life-theme level, with 79.0\%-88.0\% macro-theme repetition across all eight models. The generated events were not identical copies, but their functions repeated: care work, project administration, material processing, writing delivery, health routines, and recurring social exchanges. A loop can look locally varied while becoming narrow in what it makes possible next. The Doubao temperature probe narrows one rival explanation: if mode-lock were only a low-temperature artifact, increasing daily generation temperature from 0.75 to 1.0 should have widened the repertoire or delayed high-repetition windows. It did neither. Category count remained 11, mean 5-day repetition stayed about 96\%, and high-repetition windows moved earlier. The result does not establish a temperature law, but it weakens the view that hotter sampling alone solves recursive life-loop convergence.

Neither pressure depends on the withheld mitigation for its evidence. The model-side component is supported by the eight-model stress test, in which every included model converged under the same loop, and by the temperature probe, in which hotter sampling redistributed weight among existing channels without widening the repertoire; as noted in the limitations, this evidence does not yet separate model tendency from canon-specific attractors. The system-side component is supported by the three-year compressed simulation, in which environment watermark shells and stale stage descriptions re-entered generation through the loop's own summaries rather than through any single model call.

The architectural implication is that open-ended persona growth needs both a divergence engine and a governance system, not a larger memory store or a random event sampler. A top-level world can be real, fictional, or historical, but individual growth is shaped by the local environment that becomes reachable and meaningful. Conditional variation supplies plausible lift by eliciting long-tail life-environment material from the model's learned life-event prior. Bounded context governance then controls how much old State, history, memory, and runtime world material may constrain each future-facing step. Random events face the opposite problem. They can break repetition, but they can also break identity unless they arrive as conditional life-environment Occurrences that are plausible enough to preserve continuity and distinct enough to change what can happen next. AutoPersonas therefore treats openness as an evidence-routing problem. Occurrences must become lived material, Observations must accumulate enough support to revise State, and revision must not follow from every isolated event.

The acceptable amount of diversity is also task-dependent. In a benchmarked task agent, broader search often appears as error because success is judged against a fixed objective, tool result, or answer distribution. Persona evolution has a different tolerance profile. There is usually no single correct next life event, and plausible deviation can be useful if later evidence decides whether it hardens or disappears. AutoPersonas can therefore bear more semantic variation than a task-completion agent, but only because the OSO loop audits that variation after generation rather than treating every candidate as progress.

This places most of the technical burden on runtime rather than on any single model call. The runtime layer is where the system decides how much past evidence may constrain the present, how much present State may constrain the future, and whether a future-facing signal becomes a lived causal fact. If this layer is treated as a shallow wrapper, the system collapses back into prompt-plus-memory design: history becomes context, context becomes gravity, and future generation completes the old attractor. In AutoPersonas, runtime is the multi-timescale substrate that carries causality across past, present, and future while keeping the production mechanisms behind that substrate private.

This also separates AutoPersonas from an action planner. A planner chooses actions under current intent; AutoPersonas introduces environment-side variation, tests whether it hardens into lived evidence, and monitors whether the life trajectory moves across dimensions. Change must also accumulate across scales. Macro-only change becomes plot cutting, micro-only change updates the diary without moving the life, and environment change without lived evidence produces drift.

AutoPersonas is an application of RSI to AI persona growth, and the scope of that application should be stated precisely. Classical recursive self-improvement concerns systems that may improve their own intelligence, design capacity, or goal-directed machinery \citep{good1966ultraintelligent,omohundro2008basic}. HyperAgents gives a recent agentic version: it addresses recursive self-improvement in task systems by collapsing the vertical task/meta hierarchy into a single editable program, then evaluating descendants through predefined tasks, metrics, and archive selection \citep{zhang2026hyperagents}. AutoPersonas keeps the recursive structure of RSI, a system improving the conditions that produce its own future, but changes the object being improved. That object is not intelligence or task performance but the persona's life trajectory: the system revises the State, evidence, relationships, and life-environment conditions that shape what the persona can become. Persona growth cannot be grounded in the same objective as benchmarked task self-improvement, because a persona is a human-like life surrogate rather than a task solver. Continual goal assignment is the wrong driver, and no benchmark can cover the open space of behavior, relationships, obligations, constraints, and life-stage change. AutoPersonas therefore preserves the horizontal persona-life-environment distinction and makes the dyad recursively revisable through the OSO loop. The relevant question is whether change remains traceable, identity continuity is preserved, and State movement changes later reachability rather than producing decorative novelty.

Bounded society simulation and open-environment persona evolution therefore motivate different machinery. In a bounded simulated society, a centralized environment layer can supply public events, encounters, outcomes, and reward signals. In a deployed life-environment engine, outward growth pressure must also come from real-world information, user relationships, memory boundaries, and underdetermined Occurrences. Day-level simulation has less room to hide than coarse summaries: it exposes whether the persona acted, whether an invitation became a meeting, whether a public event entered the private life-environment, whether weak signals accumulated, and whether a relationship changed function. Shared information is projected through persona-specific State, rhythm, unresolved routes, and life-environment radius, producing a private environment that remains grounded without becoming identical for every persona. Social structure remains the most fragile part of this boundary, since a recurring person mention becomes a relationship only when it changes obligations, conflicts, mutual expectations, or later reachability.

The evidence also strengthens the generality claim. AutoPersonas is designed to be parameterized by different persona canons, life rhythms, macro-world settings, and interaction boundaries. The current evidence now goes beyond architectural compatibility: a contemporary student trajectory and a juvenile-goblin invented-world trajectory produced the same anti-fixation regime under the redesigned runtime, despite a large shift in age, species, world rules, environment radius, and contamination constraints. This supports cross-setting generality for the mechanism tested here. It does not establish statistical universality across many personas, so multi-persona long-run audits remain future work.

\section{Ethics, Safety, and Reproducibility Boundary}
\label{sec:disclosure-boundary}

Long-term persona systems may interact with real users over time. User-specific memory should remain separate from persona self-state. A persona may develop different relationships with different users, but one user's private history should not become global persona memory.

Public evaluation should not use identifiable user data. Long-run audits should use synthetic personas, internal test personas, or anonymized compressed event chains. Public examples should remove user identifiers, private relationship details, and traces that could reconstruct a conversation partner.

Generated backstory should be treated as persona-internal continuity, not as factual human biography. A system may enrich a persona's remembered past for continuity and interaction quality, but this does not imply that generated memories correspond to real human experiences.

This paper discloses causal architecture rather than production recipe. We disclose the problem definition, causal loop, subsystem roles, runtime-layer functional responsibilities, audit categories, compressed case chains, and figure-level architecture. The runtime terms used in the paper are public functional roles, not production component names. We do not disclose production internals, private logs, operational parameters, proprietary context assembly rules, internal ranking details, or implementation artifacts that would compromise privacy or commercial deployment.

The divergence configuration evaluated in Section~\ref{sec:anti-fixation-validation} is withheld in full, and this is a deliberate choice. The core contribution of this paper is the problem: a definition of self-locking, its decomposition into model-side and system-side pressures, a failure taxonomy, and measurement instruments that any long-term persona system can apply. The mitigation is reported so that the problem does not stand only as a negative result. It shows, under controlled conditions, that self-locking can be moved by runtime design. Its mechanisms belong to a commercial persona infrastructure and are not released. We recognize that this prevents independent verification of the mitigation. It does not prevent verification of the problem: the evaluator, baselines, and per-model outputs behind the self-locking measurements are public.

The disclosure boundary is:

\begin{center}
\small
\begin{longtable}{@{}p{0.22\linewidth}p{0.37\linewidth}p{0.35\linewidth}@{}}
\toprule
Disclosed & Abstracted & Withheld \\
\midrule
Self-locking definition, Observation-State-Occurrence causal loop, module responsibilities, runtime-layer causal roles, audit questions, failure taxonomy, compressed case chains, public-safe figures. & State layers, memory lanes, periodic review, external information grounding, context-governance roles, high-level evidence flow, qualitative cost classes. & Divergence-configuration internals and parameters, full prompts, named internal runtime submodules, production data schemas, private selection and sampling mechanisms, gate policies, seed-generation mechanisms, monitoring and reflection chains, thresholds, schedules, ranking weights, retry policies, token budgets, raw traces, private logs, internal paths, and user data. \\
\bottomrule
\end{longtable}
\normalsize
\end{center}

This boundary is necessary because the system is both a research artifact and a commercial persona infrastructure. The paper is intended to make the causal architecture and audit method intelligible, not to publish a deployable reproduction recipe.

\section{Data and Code Availability}

The quantitative mode-lock evidence is supported by public-safe aggregate artifacts rather than production logs. The arXiv package includes these materials as ancillary files under \path{anc/supplement/}. The eight-model stress-test materials include the aggregate manifest, per-model action-only CSVs, model-level summary CSV/JSON files, top-category summaries, strictness-variant cross-checks, the direct-loop macro-theme summary, and semantic interpretation notes under \path{anc/supplement/eight_model_action_repetition/}. The Doubao temperature probe materials include the comparison report, manifest, action-only events, derived event-level CSV for that probe, and model-level action-repetition metrics under \path{anc/supplement/doubao_temperature_probe/}. The qualitative diagnostic claims are indexed in \path{anc/supplement/qualitative_case_chains.md}, which maps each compressed case claim to the public-safe evidence type and boundary.

The action-only repetition evaluator is included as \path{anc/supplement/code/action_repetition_eval.py}. It exports daily events from \path{days.jsonl}, applies a rule-based action taxonomy over the title and action fields, and computes rolling 5-day repetition against prior history. The divergence configuration, production AutoPersonas prompts, private schemas, ranking mechanisms, operational schedules, raw private logs, and user data are outside the public release boundary.

\section{Cost and Scalability}

AutoPersonas separates response-time interaction from background evolution. Not every user message triggers full persona self-evolution. Fast loops handle local events and short-term continuity. Slower loops integrate evidence into State when enough signal has accumulated. This amortizes long-term evolution over time while keeping ordinary interactions responsive.

We report cost only qualitatively. The normalized cost structure distinguishes response-time assembly, daily life updates, periodic evidence review, monthly state review, long-horizon review, and asynchronous cultural intake. Exact production parameters are outside the disclosure boundary.

\begin{center}
\small
\begin{longtable}{@{}p{0.18\linewidth}p{0.24\linewidth}p{0.28\linewidth}p{0.24\linewidth}@{}}
\toprule
Process & Relative frequency & Relative cost & Why it exists \\
\midrule
Response-time assembly & Per interaction & Medium & Compose relevant self-state, memory, user context, and expression context. \\
Daily life update & Fast background loop & Medium & Capture lived events and short-range continuity. \\
Periodic evidence review & Slower background loop & Medium to high & Accumulate weak signals and detect repeated patterns. \\
Monthly state review & Slower background loop & High & Revise portrait-level State, role, life-environment, and life phase. \\
Long-horizon review & Infrequent background loop & High & Refresh mainline continuity and long-term possibility space. \\
Cultural intake & Asynchronous & Variable & Keep world and cultural context aligned without blocking interaction. \\
\bottomrule
\end{longtable}
\normalsize
\end{center}

This structure reflects a design choice. Long-term persona evolution should not be performed as a full self-rewrite at every turn. It should be amortized across time scales so that fast interaction and slower state revision remain separable.

\section{Limitations}

This work has eleven main limitations.

First, AutoPersonas is a systems architecture and diagnostic audit method, not a foundation-model training result.

Second, the three-year compressed simulation is diagnostic evidence. It reveals long-run failure modes and motivates the architecture, but it does not prove that open-ended self-evolution is solved.

Third, the positive validation is strongest for the central student trajectory and the juvenile-goblin fictional-world stress case. The latter provides measured cross-setting support, but the paper does not yet report variance bands over many independently sampled persona canons.

Fourth, the present evidence quantifies action-channel convergence in direct baseline persona loops, measured reversal under the redesigned runtime, and cross-world preservation of the anti-fixation regime. It does not yet quantify coverage, diversity, or fidelity across a broad set of persona canons, macro-world settings, and real-world information streams.

Fifth, unlike capability-level RSI or self-improving task-agent systems with benchmark scores and archive selection, this paper does not yet provide a scalar metric for persona-life-environment co-evolution. The action-channel and macro-theme metrics quantify baseline mode-lock symptoms, not overall life growth. HyperAgents-style evaluation can rank generated descendants on predefined tasks. AutoPersonas instead audits horizontal recursive self-evolution among Occurrences, Observations, semantic State, and life-environment possibility space. The diagnostic audit identifies failure modes and one positive trajectory, but it is not equivalent to benchmarked self-improvement evidence.

Sixth, the mitigation evidence is black-box. The paper reports controlled before/after comparisons of a withheld divergence configuration, in an isolated generator and inside the full runtime, but it neither decomposes nor releases that configuration, so the effect cannot be reproduced from the paper alone. The paper also does not include an exhaustive component-level ablation of every AutoPersonas subsystem. Future work should compare variants that remove multi-timescale revision, separate memory lanes, occurrence hardening, daily real-world information surfaces, and persona-conditioned life-environment projection.

Seventh, the eight-model action- and theme-channel stress evidence uses one complex persona canon, one direct self-orchestration loop setting, a rule-based action taxonomy, and a semantic macro-theme keeper. It supports rapid behavioral-channel and life-theme mode-lock across the included models, but it does not separate model-level tendency from canon-specific attractors, prompt design, taxonomy granularity, or keeper grouping decisions. The baseline daily life-generation temperature was 0.75, except for Kimi K2.6, whose route was code-forced to temperature 1.0; weekly and monthly compaction used temperature 0.2, and the legacy GPT judge used temperature 0. The Doubao temperature probe is also a single-model, single-canon, single-rerun comparison; weekly and monthly compaction temperature remained fixed at 0.2, so it isolates daily generation temperature rather than the whole memory pipeline. The Kimi run also used a mixed route across OpenRouter and Moonshot native access, so its model-level result should be read as a Kimi K2.6 trajectory rather than as a clean provider-route comparison.

Eighth, relationship persistence remains difficult. Events and opportunities can be generated more easily than durable social structures. Future audits should measure whether recurring people become functional relationships with mutual obligations and changed reachability.

Ninth, the attractor, gravity, and temporal-authority language in this paper is diagnostic and architectural rather than a formal mathematical proof. Formalizing self-locking as a fixed-point or attractor-basin problem, and defining entropy or reachability measures for persona-life-environment trajectories, remains future work.

Tenth, this paper does not expand the response-time social-orchestration problem. Conversation-time context selection, expression planning, and user-specific interaction strategy require separate treatment.

Eleventh, commercial and privacy constraints limit reproducibility. This paper publishes the problem definition, the causal architecture, and the diagnostic instruments; the self-locking mitigation and other production mechanisms are withheld.

\section{Conclusion}

A role-playing persona that does not grow leaves a user's relationship with it nothing to accumulate on. Long-term persona agents therefore need more than memory. They need a way to keep living in an environment that does not collapse with them. AutoPersonas addresses this problem by applying recursive self-improvement to persona growth, through a multi-timescale life-environment engine organized around controlled divergence and temporal authority separation.

Its State layer is a semantic State machine: finite enough to inspect, but open enough to represent gradual, sparse, and unexpected life change through meaning-level evidence accumulation. Environment-side Occurrences are conditioned on persona canon, current State, time scale, user relationship, macro-world signals, and real-world information, then audited for downstream State and possibility-space effects. The current evidence supports a bounded systems claim: persona-environment self-locking can be observed, categorized, and partly quantified as action-channel and theme-channel convergence, and can be mitigated architecturally by separating the source of plausible divergence from the governance of evidence, current State, future-facing Occurrences, and the revisable life-environment layer between macro-world and persona. The work provides a problem definition, a causal architecture, a quantitative mode-lock stress test, and a diagnostic audit method for long-term open-persona agents.

\section*{Acknowledgements}

The author thanks Letian Wang for the enthusiasm and vision he brought to the idea of building charming personas. His game background and sustained interest in expressive, engaging persona experiences were a source of encouragement during this work.

\section*{Supporting Materials}

\begin{itemize}[leftmargin=*]
  \item \path{main.tex}
  \item \path{references.bib}
  \item \path{figures/}
  \item \path{anc/supplement/README.md}
  \item \path{anc/supplement/code/action_repetition_eval.py}
  \item \path{anc/supplement/eight_model_action_repetition/}
  \item \path{anc/supplement/doubao_temperature_probe/}
  \item \path{anc/supplement/qualitative_case_chains.md}
  \item \path{anc/supplement/figures/}
\end{itemize}

\bibliographystyle{plainnat}
\bibliography{references}

\end{document}